\documentclass[a4paper,fleqn]{cas-sc}

\usepackage[authoryear]{natbib}
\usepackage{amsmath}
\usepackage{graphicx}
\usepackage{subcaption}

\usepackage[pagewise]{lineno}

\allowdisplaybreaks[1]
\def\tsc#1{\csdef{#1}{\textsc{\lowercase{#1}}\xspace}}
\tsc{WGM}
\tsc{QE}
\tsc{EP}
\tsc{PMS}
\tsc{BEC}
\tsc{DE}

\begin{document}
\let\WriteBookmarks\relax
\def\floatpagepagefraction{1}
\def\textpagefraction{.001}



\shorttitle{Transformer-based Map Matching Model with Limited Ground-Truth Data using Transfer-Learning Approach}
\shortauthors{Z. Jin, J. Kim, H. Yeo, and S. Choi}

\title [mode = title]{Transformer-based Map Matching Model with Limited Ground-Truth Data using Transfer-Learning Approach}


\fntext[1]{The authors appreciate Seoul Metropolitan Government and Dr. Min Ju Park for providing the taxi DTG data of Seoul city.}

\author[1]{Zhixiong Jin}[orcid=0000-0002-1370-781X]
\ead{iziz56@kaist.ac.kr}
\credit{Conceptualization of this study, Methodology, Software, Validation, Formal analysis,  Writing - original draft}

\author[2]{Jiwon Kim}[orcid=0000-0001-6380-3001]
\ead{jiwon.kim@uq.edu.au}
\credit{Conceptualization of this study, Writing - review and editing.}

\author[1]{Hwasoo Yeo}[orcid=0000-0002-2684-0978]
\ead{hwasoo@kaist.ac.kr}
\credit{Data curation, Methodology, Resources, Supervision, Funding acquisition, Project administration, Writing - review and editing}

\address[1]{Department of Civil and Environmental Engineering, Korea Advanced Institute of Science and Technology, 291 Daehak-ro, Yuseong-gu, Daejeon, Republic of Korea}
\address[2]{School of Civil Engineering, The University of Queensland, Brisbane St Lucia, Queensland, Australia}

\author[1]{Seongjin Choi}[orcid=0000-0001-7140-537X]
\cormark[1]
\ead{benchoi93@kaist.ac.kr}
\credit{Conceptualization of this study, Methodology, Formal analysis, Supervision, Writing - review and editing.}

\cortext[cor1]{Corresponding author}










\begin{abstract}
In many spatial trajectory-based applications, it is necessary to map raw trajectory data points onto road networks in digital maps, which is commonly referred to as a map-matching process. While most previous map-matching methods have focused on using rule-based algorithms to deal with the map-matching problems, in this paper, we consider the map-matching task from the data-driven perspective, proposing a deep learning-based map-matching model. We build a Transformer-based map-matching model with a transfer learning approach. We generate trajectory data to pre-train the Transformer model and then fine-tune the model with a limited number of ground-truth data to minimize the model development cost and reduce the real-to-virtual gaps. Three metrics (Average Hamming Distance, F-score, and BLEU) at two levels (point and segment level) are used to evaluate the model performance. The model is tested with real-world datasets, and the results show that the proposed map-matching model outperforms other existing map-matching models. We also analyze the matching mechanisms of the Transformer in the map-matching process, which helps to interpret the input data internal correlation and external relation between input data and matching results. In addition, the proposed model shows the possibility of using generated trajectories to solve the map-matching problems in the limited ground-truth data environment.

\end{abstract}

\begin{graphicalabstract}
\includegraphics[width=\textwidth]{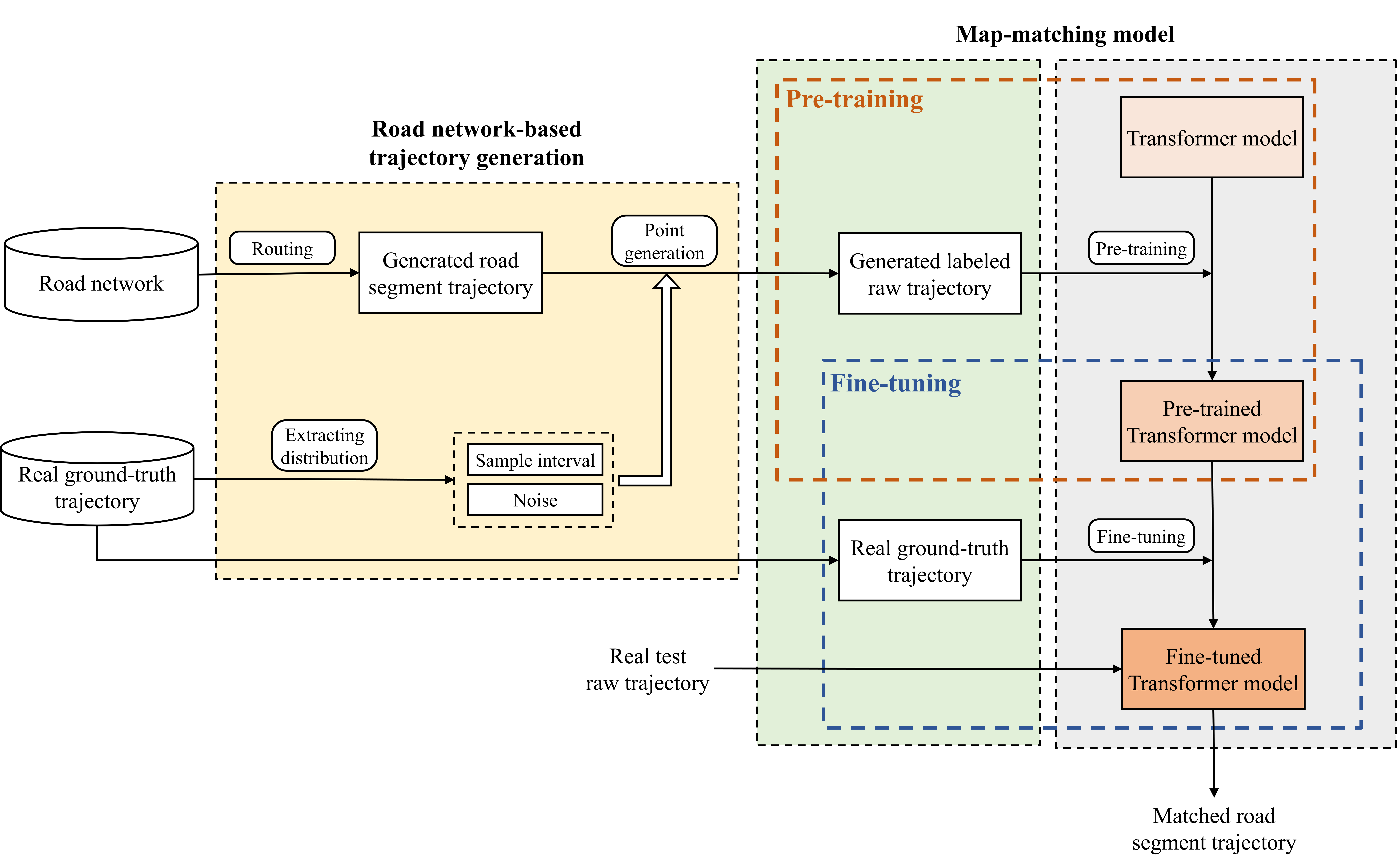}
\end{graphicalabstract}

\begin{highlights}
\item Developing a high-performing map-matching model based on Transformer.
\item A training approach based on Transfer learning to solve the limited ground-truth data problem.   
\item Interpretations of matching results based on attention mechanisms. 

\end{highlights}



\maketitle

\section{Introduction}

The proliferation of mobile devices equipped with Global Positioning System (GPS) has promoted the generation of massive amounts of GPS trajectory data, which capture user-specific mobility characteristics and system-wide spatio-temporal traffic patterns \citep{kim2015spatial,gong2017antmapper}. The big trajectory data facilitate the emergence of many trajectory-based applications such as path discovery \citep{chen2011discovering}, location/destination prediction \citep{choi2018network}, movement pattern analysis \citep{renso2013you, choi2021trajgail}, and urban planning \citep{huang2015trajgraph}. However, spatial discrepancies between recorded GPS locations and real object positions are prevalent, which raises the challenges of using GPS trajectory data in trajectory-based applications. For instance, different types of failures, such as limited satellite visibility, reflected satellite signals, and GPS receiver malfunctions, can add errors in position coordinates in the range of $5-20m$ \citep{sharath2019dynamic}. Deficiencies in current commercial digital maps can add further matching errors ranging $5-20m$ \citep{toledo2009lane}. As a result, a preprocessing procedure known as \textit{Map Matching} is necessary to correctly identify the true road segments that the moving object of a given raw GPS trajectory traveled on. \citep{quddus2007current}.

Map-matching algorithms have been studied for more than two decades to support various trajectory-based applications \citep{kubicka2018comparative,chao2020survey,hashemi2014critical}. 
%
The map-matching methods can be divided into \textit{online matching} that deals with streaming GPS data in real-time and \textit{offline matching} that processes historical trajectory data in off-line settings \citep{gong2017antmapper}. 
This paper focuses on offline map-matching, where our goal is to improve the map-matching accuracy based on historical GPS vehicle trajectory data collected from urban road networks. 
Typical offline methods include geometric algorithm \citep{bernstein1996introduction}, weight-based method \citep{sharath2019dynamic}, Kalman Filter \citep{jo1996map}, Hidden Markov Model (HMM) \citep{newson2009hidden}, and fuzzy control theory \citep{kim1998development}. Most of the existing offline map-matching algorithms take rule-based approaches, where algorithms apply pre-defined rules to process each trajectory separately to find its best matching road sequence. While these methods are fast and intuitive, processing individual trajectories independently often leads to poor map-matching performance as it cannot capture real-life, collective travel patterns embedded in a large amount of trajectory data.
%
The big trajectory data contain valuable information which helps us to understand user mobility patterns and noise characteristics. On the one hand, it is possible to leverage mobility patterns in historical trajectories to improve matching performance. The reason for this is that travel patterns between certain locations are usually highly skewed and similar trajectories can often complement each other to make themselves more complete \citep{zheng2012reducing,lin2021survey}. On the other hand, the big trajectory data can be used to extract the characteristics of GPS noise in order to further increase map-matching performance. A certain user's accumulated trajectory can disclose the position devices' noise characteristics \citep{feng2020deepmm,wang2011challenges}. Also, the aggregated trajectories gathered from various vehicles driving through a similar road network can reveal its noise characteristics induced by dense urban canyons \citep{mohamed2016accurate}, complicated road geometries \citep{merry2019smartphone}, and varying weather conditions \citep{kos2013smartphone}.   

In recent years, deep learning methods have gained popularity as powerful techniques for extracting information from big data and have achieved great successes in many fields such as natural language processing \citep{young2018recent}, computer vision \citep{voulodimos2018deep}, and speech recognition \citep{amodei2016deep}. In transportation engineering, deep learning methods are widely used in analyzing spatio-temporal characteristics of traffic to support applications such as trajectory prediction \citep{choi2018network,choi2019attention,sun2021joint}, traffic flow prediction \citep{lv2014traffic,wu2018hybrid,wang2016traffic}, and traffic signal control \citep{genders2016using,yoon2020design, yoon2021transferable}.  
In the context of map-matching, several studies have shown the possibility of developing deep learning-based map-matching models that can leverage information in big trajectory data to improve map-matching performance \citep{feng2020deepmm,zhao2019deepmm}.

However, there are two challenges for developing map-matching models based on deep learning.  
One challenge is related to the limitation of the sequential learning models that are commonly adopted as a map-matching model. Map-matching tasks can be considered solving sequence-to-sequence (seq2seq) problems, where input sequences (raw GPS trajectories) are converted into another domain of output sequences (road segments). The existing studies apply sequential learning structures based on Recurrent Neural Network (RNN) to solve map-matching problems. However, RNN has limited capability to fully capture the intercorrelation among data points in input trajectory sequences \citep{vaswani2017attention} and this can produce poor map-matching results.
Another challenge is the lack of \textit{labeled data}---ground-truth map-matched road segments for input trajectories---for training the model. Collecting a great number of labeled data for model training is often expensive and laborious \citep{kortylewski2018training}. One of the approaches to solving the problem is the data augmentation method, which artificially inflates the dataset by using label preserving transformations to add additional invariant examples \citep{taylor2018improving}. However, developing the model with generated data is inadequate due to the existence of real-to-virtual gaps, which increase the difficulty of using generated data to train the model directly. 

In order to address these two challenges, this study develops a \textit{Transformer}-based map-matching model combined with a training approach based on \textit{transfer-learning}.
To solve the problems of traditional RNN-based sequential learning models, this study uses \textit{Transformer}, a prominent deep learning model that has been successfully applied in seq2seq problems \citep{vaswani2017attention}. The Transformer is designed to consider the internal correlation of GPS points in the trajectory as well as the external relationship between input trajectory and output route, and to achieve parallel processing by using self and multi-head \textit{attention} mechanisms.  
The training approach based on transfer-learning is used to solve data sparsity problem for training. We first pre-train the Transformer model by using a large number of trajectories generated based on road network information. Then, a limited amount of available ground-truth data are used to fine-tune the model to reduce the real-to-virtual gaps. The transfer learning approach has been applied widely in computer vision to construct high-performance deep learning models since it can reduce model development costs \citep{kortylewski2018training,tremblay2018training,namozov2018efficient}.     

This paper is organized as follows. Section~\ref{sec:methodology} describes the methodology of the proposed model. In Section~\ref{sec:preliminaries}, the preliminary definitions are described. Section~\ref{sec:model} presents the model input, output and architecture. Section~\ref{sec:performance evaluation} describes the performance comparison between proposed and baseline models. Section~\ref{sec:data description} introduces the data used in this paper, and in Section~\ref{sec:baseline}, baseline models are described for model performance comparison. Section ~\ref{sec:metrics}, three metrics at different levels are introduced. In Section~\ref{sec:result}, the evaluation results are presented with proposed metrics. Finally, in Section~\ref{sec:conclusion}, conclusions and future works are presented.

\section{Methodology} \label{sec:methodology}
\subsection{Preliminaries} \label{sec:preliminaries}

In this subsection, the terms, symbols and definitions used in this paper are introduced.

\textbf{Definition 1} (GPS trajectory): A GPS trajectory $Tr$ is a sequence of chronologically ordered GPS points $Tr: p_{1} \rightarrow p_{2} \rightarrow ... \rightarrow p_{n}$. Each point $p_{i}$ has information on its GPS coordinates $(longitude, latitude)_{i}$ and timestamp $t_{i}$. Optionally, speed and heading information can be added. In this research, we only require chronologically ordered GPS coordinates without timestamp or other information. 

\textbf{Definition 2} (Road network): A road network (also called as map) is represented by a directed graph $G =(V,E)$, where a vertex $v=(x,y) \in\ V$ represents an intersection or a road end point, and edge $ e=(id,start,end,l) \in\ E$ is a directed road that starts from vertex $start$ to vertex $end$ along polyline $l$ with unique $id$.

\textbf{Definition 3} (Point-level route): A point-level route $R_{P}$ is a sequence of matched road segments, $i.e., R_{P}: e_{1} \rightarrow e_2 \rightarrow...\rightarrow e_n $ where $e_i \in E, 1\leq {i} \le {n}$. The length of matched segment is same as the input trajectory length, $i.e., length(Tr) = length(R_{P})$.

\textbf{Definition 4} (Segment-level route): A segment-level route $R_{S}$ is expressed as a sequence of connected road segments, $i.e., R: e_{1} \rightarrow e_2 \rightarrow...\rightarrow e_m $ where $e_i \in E, 1\leq {i} \le {m},m \le {n} $, and $e_i.end = e_{i+1}.start$. The length of route is less than input trajectory's $(Tr)$ length, $i.e., length(R_{S}) \leq length(Tr)$.

\textbf{Definition 5} (Map matching): Map-matching $ {MM}_{G}:Tr \rightarrow R_{P/S} $ is the process of finding the point or segment-level route $R_{P/S}$ based on the GPS trajectory $Tr$ in a given road network $G$.
%
%
In other words, map-matching is the process of converting input GPS trajectories into the corresponding road segments.



\subsection{Map-matching Model} \label{sec:model}

We develop a Transformer-based map-matching model and propose a training approach based on transfer learning in this study. 
The proposed map-matching method consists of two main approaches: $(1)$ Transformer-based map-matching model development and $(2)$ Transfer-learning approach for model training with limited ground-truth data. 
In Section~\ref{sec:Transformer}, we explain how the Transformer is used in our map-matching problem in detail. In Section~\ref{sec:transfer learning}, we first generate various training datasets with different parameters for model pre-training. Then, the limited ground-truth data is used for fine-tuning the pre-trained Transformer-based map-matching model to improve matching performance and reduce the real-to-virtual gaps that exist between generated and real-world trajectories. In the following sections, we introduce each approach in detail.

\subsubsection{Transformer-based Map-matching Model} \label{sec:Transformer}
\begin{figure}[!ht]
  \centering
  \includegraphics[width=1\textwidth]{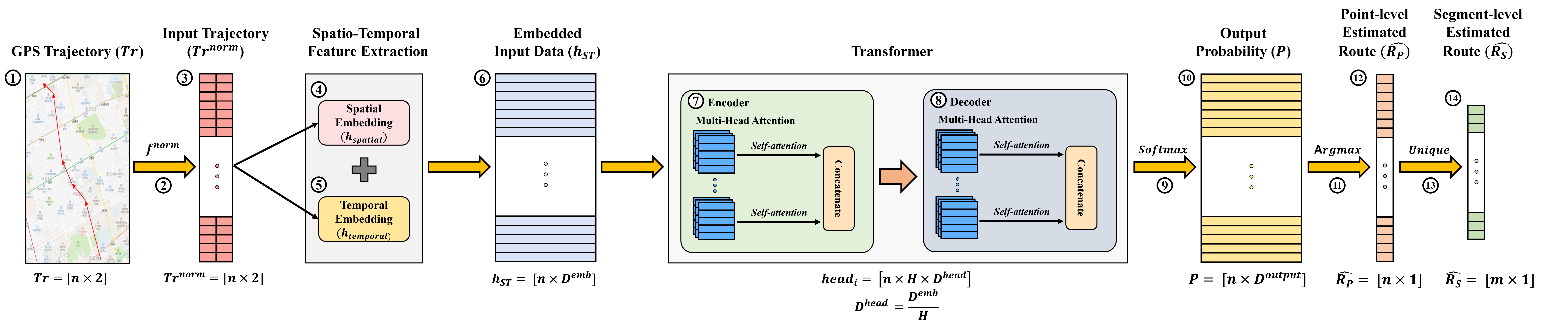}
  \caption{Transformer-base Map matching process }\label{fig:map_matching_process}
\end{figure}

In this section, we explain the input and output structure of the proposed map-matching model. Then, spatio-temporal feature extraction method of input GPS trajectory is introduced. Finally, the architecture of map-matching model is presented. The Transformer-base map-matching process is shown in Figure~\ref{fig:map_matching_process}

\paragraph{\textbf{Model Input and Output}}
As discussed earlier, the input of map-matching model is the GPS trajectory ($Tr$) and the output is the point or segment-level route ($R_{P/S}$). Each GPS trajectory contains $n$ GPS points.
\begin{equation}
  \begin{split}
  Tr = [p_1,\cdots, p_n ] = \Big[\big(lat_1, long_1\big),\cdots, \big(lat_n, long_n \big) \Big]
  \end{split}
\end{equation}

Instead of using raw GPS points, in this study, we use a normalized GPS trajectory to make the training faster and reduce the possibility to get stuck in local optimal solution \citep{sola1997importance}. The normalized GPS trajectory is denoted as $Tr^{norm}$ as shown in Eq.\ref{eq:norm}.
\begin{equation}\label{eq:norm}
  \begin{split}
  & Tr^{norm} = f^{norm} (Tr) = \Big[ \big(f^{norm}(lat_1) , f^{norm}(long_1)\big) , \cdots , \big(f^{norm}(lat_n) , f^{norm}(long_n) \big) \Big]
  \end{split}
\end{equation}

The normalization function is defined as 
\begin{equation}
  \begin{split}
  &f^{norm}(X) = \frac{X-X_{min}}{{X_{max}-X_{min}}}\\
  \end{split}
\end{equation}

where $X$ represents the GPS coordinate longitude or latitude, $X_{max} and X_{min}$ are the maximum and minimum longitude or latitude values in the target network. In Figure~\ref{fig:map_matching_process}, steps 1-3 present the normalization process from GPS trajectory ($Tr$) to input trajectory ($Tr^{norm}$). 

The first output of the proposed model is the  point-level estimated route $\hat{R_{P}}$, which contains $n$ matched edges $\hat{e_{n}}$  for each GPS point. In Figure~\ref{fig:map_matching_process}, steps 3-12 shows the map-matching process from input trajectory ($Tr^{norm}$) to  point-level estimated route ($\hat{R_{P}}$).
%
\begin{equation}
  \begin{split}
    & \hat{R_{P}} = {MM}_G (Tr^{norm}) = [\hat{e_1}, \cdots, \hat{e_n}] 
  \end{split}
\end{equation}

To further obtain the segment-level estimated route $\hat{R_{S}}$, the unique values $\hat{e_{m}}$ are chosen in the point-level estimated route $\hat{R_{P}}$ without sorting since the order of the value provides the vehicle's traveling direction information. In Figure~\ref{fig:map_matching_process}, steps 12-14 show the process of obtaining $\hat{R_{S}}$ from $\hat{R_{P}}$.

\begin{equation}
  \begin{split}
    & \hat{R_{S}} = Unique(\hat{R_{P}}) = [\hat{e_1}, \cdots, \hat{e_m}] 
  \end{split}
\end{equation}
\noindent



\paragraph{\textbf{Spatio-Temporal Feature Extraction}}

The GPS trajectory is one of the spatio-temoral data since it includes both spatial and temporal information. In map matching, the spatial information of input trajectory is used to detect the location of the GPS points and find noise distribution, while the temporal information helps us comprehend the order of the GPS points, indicating the moving direction of the vehicle. As a result, it is crucial to extract both features properly in the map-matching process. In this research, learned positional embedding \citep{gehring2017convolutional} is applied for temporal information extraction, while a novel GPS embedding method  is used to extract spatial features of the input trajectory.

\begin{itemize}
    \item \textbf{Spatial Feature Extraction}\\
In deep learning, it is common to use feature-extraction layers to convert input variables into feature vectors. In map matching, it is necessary to extract the spatial features from GPS points to properly train the deep learning model. Previous researches that used deep learning for map-matching problems discretized the road networks into zones and used embedding or one-hot encoding to extract the spatial features \citep{zhao2019deepmm,feng2020deepmm}. However, using discretized road network is ineffective in map-matching tasks due to the following reasons.
First, the noise characteristics of GPS points are neglected. As stated earlier, extracting the noise characteristics from accumulated trajectory data is one of the important approaches to reduce noise effects in the map-matching process. It is hard to achieve map-matching task at segment level without considering GPS noise distribution properly since if we discretize the road networks into several zones, there might be several segments in a certain zone, which increase the uncertainties in the map-matching procedure. 
Second, the relationships between GPS points in a trajectory are ignored. There are close relationships between GPS points due to GPS points' continuity in the trajectory. These relationships can help improve the matching performance of the map-matching model, which can not be neglect.  
As a result, in this study, we use multiple fully connected layers to properly extract the spatial features of GPS points without losing important information. 

\begin{equation}\label{eq:spatial}
  \begin{split}
  & h_{spatial} = FC( Tr^{norm} ) \\
  \end{split}
\end{equation}

\item \textbf{Temporal Feature Extraction}\\
In the Transformer, additional positional representation, also known as \textit{positional embedding} is required to model the temporal features of input GPS data since the positional information of input data is ignorant. In this research, the positional embedding is defined as, 
\begin{equation}\label{eq:positional embedding}
  \begin{split}
  & h_{temporal} = Embedding(Arrange(len(Tr^{norm})) \\
  \end{split}
\end{equation}
\noindent
where $Embedding$ is the dense representation that converts discrete position feature to continuous vector form, $Arrange(X)$ generates the integer number from 0 to $X$-1, and $len(Tr^{norm})$ represents the length of trajectory. 
\cite{vaswani2017attention} demonstrates that the results are similar between cosine function-based positional encoding and learned positional embedding which is used in this research.

\end{itemize}

After extracting spatial features from Eq. \ref{eq:spatial} and temporal features from Eq. \ref{eq:positional embedding}, we combine these two features to get a spatio-temporal feature representation of input data. The corresponding expression is shown as,

\begin{equation}\label{eq:spatio_temporal embedding}
  \begin{split}
  & h_{ST} = h_{spatial} + h_{temporal} \\
  \end{split}
\end{equation}
where both of them have same dimension $D^{emb}$, which is dimension of embedding. In Figure~\ref{fig:map_matching_process}, steps 3-6 present the process of spatio-temporal feature extraction of input trajectory($Tr^{norm}$).

\paragraph{\textbf{Transformer Architecture}}

In the proposed map-matching model, there are two reasons for choosing Transformer to deal with map-matching problems. 
The Transformer can capture the internal correlation of the GPS trajectory and the external correlation between the GPS trajectory and the output route (or segment-level route $R_{S}$ ). The correlations are primarily captured by \textit{attention modules} in both encoder and decoder \citep{lu2021pretrained}. In the encoding stage, the attention modules capture the internal correlation of GPS trajectory, which helps to understand the relationship between each GPS points in the trajectory and reduces the effects of uncorrelated GPS points in map-matching processes. Similarly, the attention modules in the decoding stage extract the relationship between the input GPS trajectory and the output route to enhance matching performance and to analyze and interpret the matching result. Therefore, the matching performance at confusing regions such as the initial segment, last segment, and the transition area between two consecutive segments can improve due to stated characteristics. 
%

\begin{figure}[!ht]
  \centering
  \includegraphics[width=0.6\textwidth]{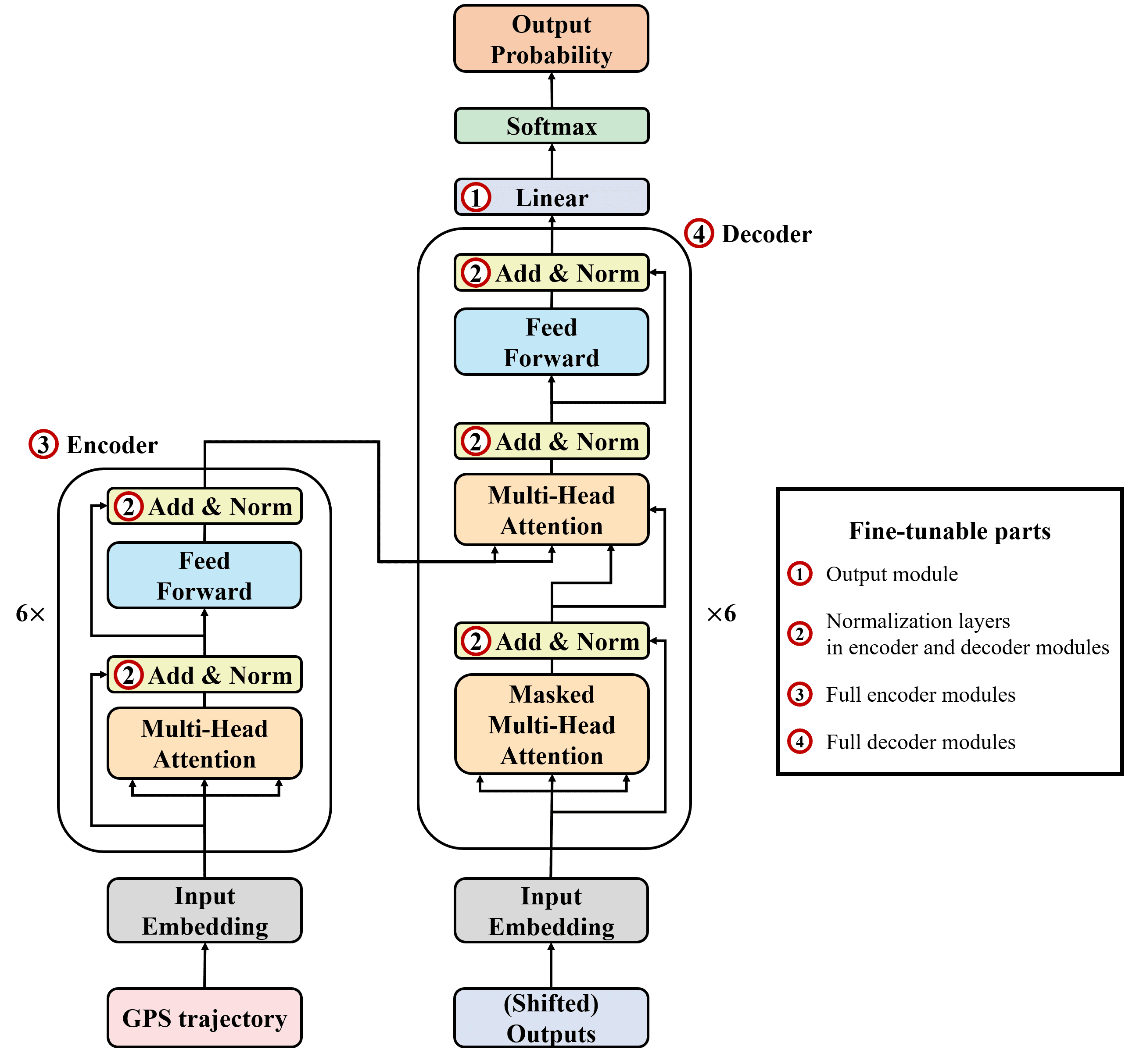}
  \caption{Overview of Transformer architecture}\label{fig:Transformer}
\end{figure}

The architecture of the Transformer is shown in Figure~\ref{fig:Transformer}. It mainly consists of four components: input embedding, encoder, decoder, and output modules. The input embedding module includes spatial and temporal embedding. It extracts the spatio-temporal features of input trajectory, which has been introduced previously. The encoder and decoder modules are mainly composed of multi-head attention, normalization layers, and position-wise feed-forward networks. The encoder module is used to understand spatio-temporal features of GPS trajectory data, and generate a representation ($h_{enc}$) for the observation sequence based on the extracted information of input data ($h_{ST}$). Conversely, the decoder module converts encoded information from the encoder module to the target output sequence. The output module also functions as a classification module, classifying the input GPS points into segments. The output module composes a fully connected layer. 
In the next, three main components in encoder and decoder modules are introduced.

\begin{itemize}
    \item \textbf{Self and Multi-head Attention}\\
In self-attention model, query matrix $Q$, key matrix $K$ and value matrix $V$ have dimensions $d=d_k=d_q=d_v$, respectively. The attention equation used in Transformer is shown as, 
 
\begin{equation}
  \begin{split}
  Attention(\mathbf{Q,K,V}) =  Softmax(\frac{\mathbf{QK}^T}{\sqrt{D_k}})\mathbf{V} = \mathbf{AV}
  \end{split}
\end{equation}

\noindent
where $\mathbf{A} = softmax(\frac{\mathbf{QK}^T}{\sqrt{D_k}})$ is attention matrix. To alleviate gradient vanishing problem of softmax function, the dot-products of queries and keys are divided by $\sqrt{D_k}$.

Instead of simply applying a single attention function, it has been found that it is beneficial to map the queries, keys, and values for $H$ times to learn different contextual information respectively. In other words, the dimension of each head $d$ is equal to $\frac{D^{emb}}{H}$, $i.e., D^{head} = d = \frac{D^{emb}}{H}$. The self-attention function is performed on each projected version of queries, keys, and values in parallel. Then the results from each self-attention function are concatenated and projected again to obtain the weight of final values. The aforementioned process is defined as multi-head attention, which is shown as, 
\begin{equation}
  \begin{split}
  &MultiHeadAttn(\mathbf{Q,K,V}) =  Concat(head_1,head_2...head_H) \mathbf{W}^O \\
  &head_i = Attention(QW_i^Q,KW_i^K,VW_i^V)
  \end{split}
\end{equation}
\noindent
where $W_i^Q,W_i^K,W_i^V,W^O$ are the projections of parameter matrices in queries, keys, values, and output, respectively.In Figure~\ref{fig:map_matching_process}, steps 7-8 presents the multi-head attention process of encoder and decoder.

\item \textbf{Normalization layer}\\
In the encoder and decoder modules, the Transformer uses a residual connection \citep{he2016deep} around multi-head attention and position-wise feed-forward network, followed by Layer Normalization (LN) \citep{xu2019understanding,ba2016layer}. LN is defined as 

\begin{equation}\label{eq:layer normalization}
  \begin{split}
  & LayerNorm(\textbf{X}) = \textbf{g} \cdot N(\textbf{X})+\textbf{b}\\
  &N(\textbf{X}) = \frac{\textbf{X}-\mu}{\sigma}\\
  &\mu =\frac{1}{D^{emb}}\sum_{i=1}^{H} x_{i}\\
  &\sigma=\sqrt{\frac{1}{D^{emb}}\sum_{i=1}^{D^{emb}}(x_{i}-\mu)^2}   \\
  \end{split}
\end{equation}
\noindent

where $\textbf{X} = (x_1, x_2 \cdots x_{emb})$ is the input vector with size $D^{emb}$.  $\mu$ and $\sigma$ are the mean and standard deviation of input. $\textbf{b}$ and $\textbf{g}$ are the trainable parameters with the same dimension $D^{emb}$. LN is considered as a mechanism, which is effective at stabilizing the hidden state dynamics in recurrent networks \citep{ba2016layer}. 

\item \textbf{Position-wise Feed-Forward Network}\\
The position-wise feed-forward network consists of two linear transformations separated by a RELU activation. Each network in the encoder and decoder modules operates separately and identically in each position. The equation is defined as,
\begin{equation}\label{eq:feed forawrd network}
  \begin{split}
  & FFN(x) = ReLU(W_1x+b_1)W_2+b_2\\
  \end{split}
\end{equation}

where $x$ is the outputs of previous layers, and $W_1,W_2,b_1,b_2$ are trainable parameters. Even though the network is simple, it is important for the Transformer to achieve good performance since it can prevent rank collapse problems, which can occur when the self-attention are simply stacked \citep{lin2021survey}.

\end{itemize}

%
%
%

After getting results from decoder modules, we use the output module to do the classification. The output module uses a fully connected to change the dimension from $D^{emb}$ to $D^{output}$, where $D^{emb}$ is the dimension of embedding and $D^{output}$ is the target output dimension of our model. In this research, $D^{output} $ = number of link +1 due to padding tokens in input trajectory. In addition, the $Softmax$ function is used to calculate the matching probability of each GPS point at each segment. The corresponding process is shown as, 

\begin{equation}\label{eq:Softmax}
  \begin{split}
  & P = Softmax(FC(h_{dec}))   \\
  \end{split}
\end{equation}

\noindent
where $h_{dec}$ represents the result from decoder modules. $FC$ is the fully connected layer, $P$ output probability and $Softmax$ represents the softmax function. The dimension of $P$ is same as input data $Tr^{norm}$. In Figure~\ref{fig:map_matching_process}, steps 8-10 depict the process of calculating output probability ($P$) from model output. 

After obtaining output probability $P$, we use function $argmax$ to get the point-level estimated route ($\hat{R_{P}}$). The corresponding expression is shown as,
 
\begin{equation}\label{eq:final step}
  \begin{split}
  & \hat{R_{P}} = argmax (P) = [\hat{e_1}, \cdots, \hat{e_n}]  \\
  \end{split}
\end{equation}
where $\hat{e_n}$ indicates the best matched segment for each GPS point. In Figure~\ref{fig:map_matching_process}, steps 10-12 depict the process of calculating the point-level estimated route ($\hat{R_{P}}$) from output probability ($P$).


%

\subsubsection{Transfer Learning Approach for Model Training with Limited Ground-Truth Data} \label{sec:transfer learning}
As stated earlier, collecting a great number of labeled data for model training is often expensive and laborious \citep{kortylewski2018training}. Therefore, one of the biggest challenges in deep-learning based map-matching development is building a high-performing model with limited ground-truth data. In this research, we combine data augmentation (or data generation) method and the transfer-learning approach to overcome the challenge and develop a high-performing map-matching model. The data augmentation method is used to generate a great number of road network-based trajectory data for model pre-training, while the limited ground-truth data is used in model fine-tuning to reduce the real-to-virtual gaps. The concepts of pre-training and fine-tuning are from transfer learning. In computer vision \citep{kortylewski2018training,tremblay2018training,namozov2018efficient}, the researchers have successfully used stated methods and have demonstrated the potential to develop the high-performing model using generated data and limited ground-truth data. In this section, we introduce how we built our model via data augmentation (data generation) method and transfer-learning approach.  

\begin{figure}[!hb]
  \centering
  \includegraphics[width=0.8\textwidth]{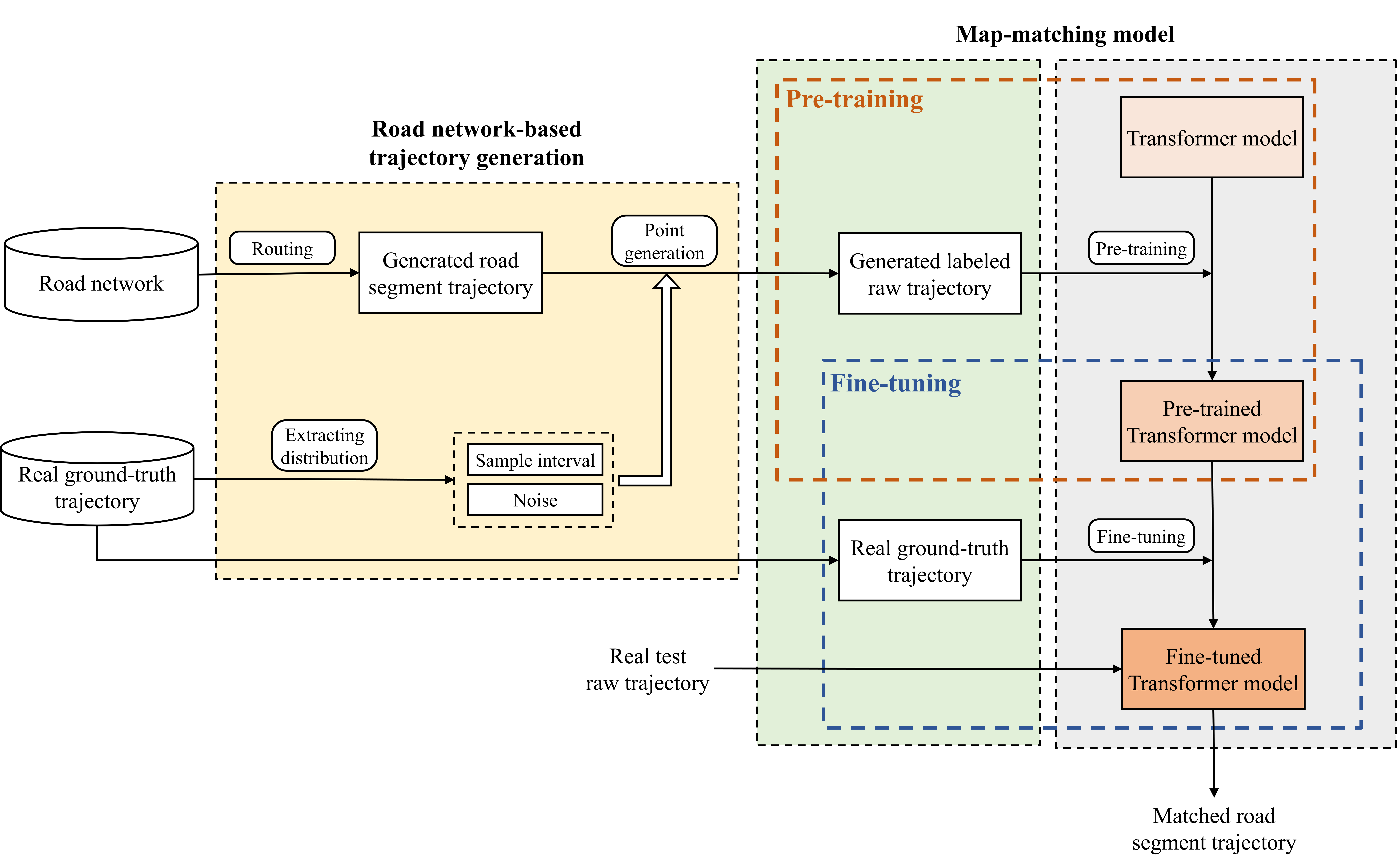}
  \caption{The framework of proposed map-matching task}\label{fig:The_framework_of_Model}
\end{figure}



\paragraph{\textbf{Pre-training with Generated Trajectory Data}}

%
In map matching, data augmentation or data generation methods are used to solve the data sparsity problem. The data augmentation methods in trajectory generation are categorized as rule-based and data-based models. The rule-based augmentation methods are defined as generating trajectories based on pre-defined rules. Most researches focus on generating trajectories based on the shortest paths between two locations since they are simple, intuitive, and fast. However, people do not always choose the shortest path in reality, for example, drivers may choose a longer path due to short travel time or inevitable situations. Therefore, it is necessary to consider various scenarios to improve the matching performance of the deep learning model. 
Different from rule-based trajectory augmentation methods, the data-based methods generate trajectories based on the characteristics of known data. Typical algorithms are data duplication \citep{travis2008trajectory}, Markov chain \citep{chen2011discovering}, Generative Adversarial Network (GAN) \citep{wang2021large}, and Generative Adversarial Imitation Learning \citep{choi2021trajgail}. However, these methods need a sufficient number of labeled data to cover the various scenarios in the target area, which is hard to apply in our map-matching task.

In this study, our goal is to develop a high-performing map-matching model with limited ground-truth data. Therefore, rule-based trajectory algorithms are more suitable since the performance of trajectory generation does not strongly rely on the size of collected data. We propose a rule-based trajectory generation method based on the road network data, which is cost-effective and can generate various scenarios.
The proposed trajectory generation method is divided into four steps: \textbf{\emph{Route Generation, Point Generation, Point Selection, and GPS Trajectory Generation}}. 

\begin{figure}[!ht]
  \centering
  \includegraphics[width=0.8\textwidth]{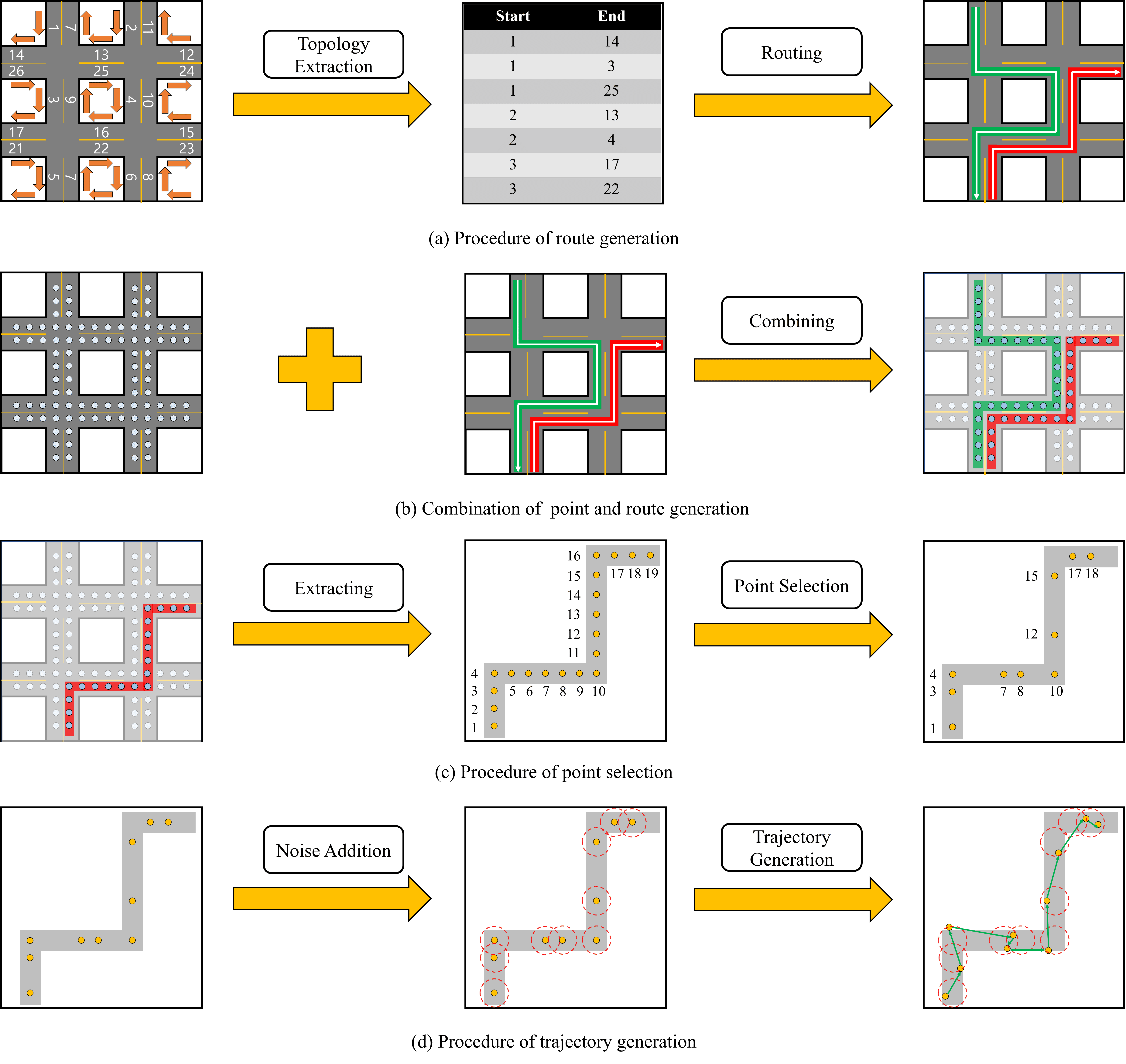}
  \caption{Overview of data generation architecture }\label{fig:data_generation}
\end{figure}

\begin{itemize}
    \item \textbf{\emph{Route Generation}}\\
First, a segment connection table is defined using topological information from the road network. The table consists of \textit{start} and \textit{end} columns that the vehicles move from segment in \textit{start}  to segment in \textit{end}. Then, all feasible routes are generated based on the constructed table. The length of the route is determined by the number of linked segments \textbf{\emph{N}}. More complex routes can be generated when \textbf{\emph{N}} increases. Figure~\ref{fig:data_generation} (a) shows the process of route generation.
%

\item \textbf{\emph{Point Generation}}\\
In the point generation step, we generate points with constant distance \textbf{\emph{D}} on the road network. The generated points are labeled with road segment ID and their original ID. Then we combine the information from point and route generation results to get an initial ground-truth GPS trajectory. The process is shown in   Figure~\ref{fig:data_generation} (b) in detail.


\item \textbf{\emph{Point Selection}}\\
After generating the initial ground-truth trajectory, the number of points at each segment should be determined. The amount of points at each segment is mainly affected by ground-truth trajectory distribution. The number of points is different at each road segment due to various sampling intervals and segment lengths. Therefore, we estimate the point selection range $[r_{1},r_{2}$ and randomly choose points inside it to guarantee that the sampling intervals of produced points are close to real trajectories' sampling intervals. Furthermore, the order of the point selection is based on the road direction that the selected point ID in the previous step cannot be greater than the selected point ID in the present step. For example, in  Figure~\ref{fig:data_generation} (c), if we choose a point with ID 4, we cannot choose points with IDs are 1,2, and 3 in the next step. The reason for this is that vehicles can only go ahead along the direction of the roads. After step 3, the final ground-truth trajectories are generated.

\item \textbf{\emph{GPS Trajectory Generation}}\\
In the final step, we generate raw trajectories for training the model based on the generated trajectories obtained in step 3. We add noise at each point on each trajectory to ensure that the generated raw trajectories are close to real-world trajectories. We calculate the Pearson's correlation \citep{ahlgren2003requirements} coefficient between the GPS noise along longitude and latitude. The correlation coefficient is $0.094$ with $0$ P-value. The result indicates that the noise along longitude and latitude are uncorrelated. In addition, the noise distribution at longitude or latitude is close to zero-mean Gaussian distribution. The means and standard deviations of the longitude and latitude noise are $0.427m$ and $15.653m$ and $-5.467e^{-6}m$ and $14.153m$, respectively. The previous research has demonstrated the point of view \citep{feng2020deepmm}. Therefore, in this research, we also assume that the spatial noise for each coordinate follows zero-mean Gaussian distribution, which is shown as, 

\begin{linenomath}
  \begin{equation}
  f(x_{long/lat}|\sigma_{noise}^2) =\frac{1}{\sqrt{2\pi\sigma_{noise}^2}} e^ \frac{-x_{long/lat}^2}{2\sigma_{noise}^2}
  \end{equation}
\end{linenomath}

\noindent where $x_{long/lat}$ denotes the spatial coordination (longitude or latitude), $\sigma_{noise}$ is the standard deviation of the Gaussian distribution. In this research, different generated raw trajectories with the same ground-truth trajectory are produced by using different $\sigma$ values. Figure \ref{fig:data_generation} (d) depicts the process of GPS trajectory generation.

In conclusion, the raw trajectory data are generated by using information from the road network and ground-truth trajectory data. After the trajectory generation step, the generated trajectories are ready for pre-training the deep learning model.

\end{itemize}
We first pre-train our Transformer model by using generated trajectories with various sampling intervals and noise distribution. In this paper, the amount of generated trajectory data used for model training is 240,000. Additionally, the pre-trained Transformer model is composed of eight attention heads and six layers for each encoder and decoder block. When pre-training the model, the loss function is defined by cross-entropy loss between the predicted output point-level route $\hat{R_{P}}$ and the ground-truth route $R_{P}$. Via backpropagation with the Adam optimizer, we train the network with a learning rate of 0.0007 and the dropout value of 0.1. 


\paragraph{\textbf{Fine-tuning with Limited Ground-Truth Data}}

Even if we try to generate data close to real trajectories as much as possible, there are still differences between the two datasets, which means that it is not enough to develop map-matching algorithms only depending on generated trajectory data. In trajectory generation, the datasets are generated by the specific noise distributions ($\sigma_{noise}$) and sampling intervals. However, these two factors can be different from real-world trajectories due to complex and changing communication environments. To fill the real-to-virtual gaps of two datasets, we choose to use fine-tuning method   
from transfer learning. In machine learning and deep learning, the term \textit{fine-tuning} is often used to describe the optimization process of hyper-parameters during the validation step. However, in the context of ``transfer learning,'' \textit{fine-tuning} refers to the process of transforming a model trained on one domain (problem) to a new domain (problem).
Depending on the fine-tuning dataset size and similarity with pre-trained dataset, there are four general scenarios to fine-tune the model  \citep{yosinski2014transferable}.

\textbf{\emph{(Scenario 1): For a large ground-truth dataset that is different from the pre-trained model's generated dataset}} It is preferable to fine-tune all of the model's layers since the fine-tuning dataset is large enough to train the model and significantly different from the pre-training dataset.

\textbf{\emph{(Scenario 2): For a large ground-truth dataset that is similar to the pre-trained model's generated dataset}} It is feasible to freeze components of the layers to fine-tune the model since the fine-tuning dataset is similar to the pre-training dataset. Even if fine-tune the whole model, there is no risk of over-fitting since the dataset is sufficient to re-train the model.

\textbf{\emph{(Scenario 3): For a small ground-truth dataset that is different from the pre-trained model's generated dataset}} It is the most difficult scenario to deal with and occurs frequently in the real fine-tuning problems. The reason is that there are significant differences between the pre-trained dataset and ground-truth dataset, which means that we cannot fine-tune for a small number of layers without taking risks of overfitting problems owing to the small amount of tuning dataset. In this case, it is necessary to fine-tune only an appropriate number of layers, which are difficult to control. 

\textbf{\emph{(Scenario 4): For a small ground-truth dataset that is similar to the pre-trained model's generated dataset}} It is one of the special cases of scenario 3. In this scenario, we can also use the fine-tuning technique from scenario 3 that choose the appropriate layer for fine-tuning. The main difference is the number of fine-tuning layers might be less than the previous scenario since the ground-truth dataset is similar to the pre-training dataset. 

The best scenario for fine-tuning the model is second one that there is a large ground-truth dataset that is similar to the pre-training dataset. Even though the ground-truth dataset for fine-tuning differs from the pre-training dataset, we retrain the entire model to build a high-performing map-matching model if the dataset is large enough. However, it is challenging and laborious to collect a large amount of ground-truth data for model training in reality. Instead, we have to use small amount of ground-truth dataset to build a high-performing model to solve the problems. Therefore, scenario 3 and 4 are the two feasible scenarios in this study since the goal is to develop a high-performing map-matching model using limited ground-truth data.

Despite the fact that both scenarios 3 and 4 require discovering appropriate layers for fine-tuning the model, the difficulties of implementing fine-tuning are different. In other words, scenario 4 is considerably easier than scenario 3 since the real-world trajectories in the former situation are close to the pre-trained model's generated dataset. As a result, having prior information of the real-world trajectories helps us in the generation of more realistic trajectories and reduces fine-tuning challenges. However, there are situations when we are unable to obtain any information of real trajectory data, making it more difficult to build a high-performing map-matching model. In other words, it is hard to generate a realistic training dataset, which increases the difficulties in the fine-tuning process. As a result, it is preferable to consider both situations and generate two different datasets for pre-training, one of which is similar to the real-world instance and the other completely different, and then use the limited ground-truth dataset for fine-tuning. Furthermore, throughout comparing the performance of two models that are pre-trained with different generated trajectory data and fine-tuned with the same ground-truth data, we can demonstrate the importance of prior knowledge of real-world trajectories and determine how much fine-tuning improves model performance.

In this study, we choose four fine-tuning components in Transformer: {\large \textcircled{\small 1}} output module, {\large \textcircled{\small 2}} normalization layers in encoder and decoder modules,{\large \textcircled{\small 3}} full encoder modules, and {\large \textcircled{\small 4}} full decoder modules. The red circled numbers in Figure~\ref{fig:Transformer} depicts the fine-tunable components. To identify the appropriate fine-tuning layers for both cases, we fine-tune each module individually first, then gradually increase the number of tuning modules. The corresponding results are introduced in Section~\ref{sec:fine-tuning evaluation} in detail.

\section{Performance Evaluation}\label{sec:performance evaluation}
\subsection{Data Description}\label{sec:data description}

\begin{figure}[!ht]
  \centering
  \includegraphics[width=0.5\textwidth]{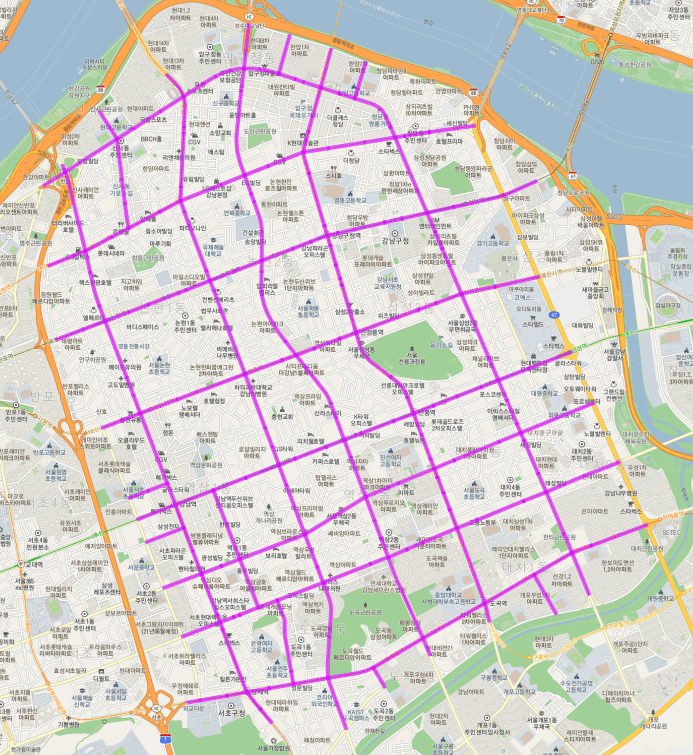}
  \caption{Road network in Gangnam district (Background: Kakao map (https://map.kakao.com))}\label{fig:Gangnam}
\end{figure}

A case study is designed to evaluate the performance of the proposed map-matching model, using data collected by digital tachographs (DTG) installed at taxis operating in Gangnam District in Seoul, South Korea.
The DTG collects information such as driving position (longitude and latitude), speed, and passenger occupancy. The collected data are converted into a taxi trajectory dataset by chronologically linking the data points of the same taxi ID. In this study, we choose the taxi trajectories that traveled in the Gangnam district where GPS errors occur frequently due to the complex urban environment situation. The Gangnam district consists of 228 major road segments in a grid structure as shown in Figure ~\ref{fig:Gangnam}. We further divide the trajectory with the same ID into several sub-trajectories, which cover the trip for each passenger, because each taxi trajectory comprises several trips associated with various passengers. We choose the passenger-level sub-trajectories, which lengths are greater than two kilometers and average sample intervals are around 20 seconds.
After the data preprocessing step, we manually label each GPS point to its corresponding segment, obtaining 1,331 vehicle trajectories with 35,127 GPS points. In our dataset, various scenarios are included such as congestion, detouring, and U-turn. In this research, we randomly select 931 trajectories from labeled ground-truth taxi trajectory data (70\% of the dataset) for training and fine-tuning the model, while the left 400 labeled passenger-level trajectories (30\% of the dataset) are used as the test dataset for evaluating models' performances.

\subsection{Baseline Models}\label{sec:baseline}

We compare the performance of our model with four baseline models.
\begin{itemize}
\item \textbf{\emph{FMM}} \citep{yang2018fast}: It provides a fast map-matching technique that is both efficient and scalable. FMM uses two techniques to incorporate the hidden Markov model: $1)$ pre-computing the upper bound origin-destination table to store all pairs of shortest routes; $2)$ using fast hash table search to replace repetitive routing queries.

\item \textbf{\emph{ST-Matching}} \citep{lou2009map}: It's widely used in GPS trajectories with low-sampled cases. It creates a candidate graph to identify the best-matched route by taking into account the road network's spatial and topological characteristics, as well as the temporal aspects of trajectories.

\item \textbf{\emph{LSTM-based seq2seq model}} \citep{zhao2019deepmm}: It is less affected by dense noise and it is also powerful in the urban area. In the seq2seq model, the input trajectories are compressed to context vectors in encoding states, and the road segment trajectories are translated at decoding states. The RNN cells used in both encoding and decoding states are LSTM. 

\item \textbf{\emph{LSTM-based attentional seq2seq model}} \citep{feng2020deepmm}: It is more efficient in the map-matching process since it solves the long-dependency problems existing in RNN-based seq2seq models. Specifically, if a fixed-size context vector is used, there may be diminishing problems when translating long raw trajectories, leading to the accuracy drop. To solve the problem, a dynamic context vector generated from encoder hidden states for each translation step is used to make an attention layer in the attentional seq2seq model.  The LSTM cells are used in both encoding and decoding states. 
\end{itemize}

\subsection{Evaluation Metrics} \label{sec:metrics}


We use three metrics for evaluation -  Average Hamming Distance (AHD), F-score, and BLEU at point and segment level to compare and evaluate the overall performance of map-matching models. The point-level matching analyzes the matching result at the point-level route($R_{P}$). The point-level matching is beneficial to find how the map-matching model matches each point accurately to its corresponding segment. Most previous researches, particularly in the field of online map-matching models, have used various point-level approaches to evaluate their map-matching model. Even though it is crucial to properly match each point, route integrity is also an essential factor to consider in map-matching tasks, especially in offline map-matching models. The route integrity refers to how completely the map-matching models match the GPS points to road segments at trajectory level. For example, suppose the ground-truth result at point-level route ($R_{P}$) is $[98,98,98,97,97,97,1,1,3,3]$, and the candidates are $[98,98,97,97,97,1,1,1,3,3]$ and $[98,98,107,97,97,1,1,183,3,3]$. Although both results have a matching accuracy of 0.8, the first one is better in route integrity than the second. As a result, not only map accuracy but also route integrity should be taken into account. 
We define segment-level matching, which focuses on matching results at the segment-level route ($R_{S}$). We select the unique value in the matching result without changing the order. In the previous example, the segment-level ground-truth result is transformed into $[98,97,1,3]$. Segment-level matching is beneficial for minimizing the impact of confusing points in map-matching results. In practice, it is difficult to annotate the points manually at some regions where there are multiple segments close to the target matching points on the actual path \citep{taguchi2018online}. Therefore, we need to analyze the result in segment-level matching. As previously stated, the ground-truth result at segment level is $[98,97,1,3]$ and the two candidates are converted as $[98,97,1,3]$ and $[98,107,97,1,183,3]$, respectively. In this situation, the former's matching accuracy is 1 and the latter's should be transformed first by using the sequence alignment method for accuracy-calculation.

Therefore, we evaluate the model performances with three metrics at point and segment level to consider the point matching accuracy and route integrity. The Average Hamming Distance (AHD) and F-score are used to calculate the accuracy and the Bilingual Evaluation Understudy (BLEU) score is useful to consider the order of result sequence and its integrity.

\paragraph{\textbf{Average Hamming Distance (AHD)}}
In information theory, the Hamming distance measures the number of positions in which the corresponding symbols are different between two equal-length strings \cite{hamming1986coding,bookstein2002generalized}. Average Hamming Distance is defined as the total number of different symbols divided by the length of string \cite{shutao1998average,zhang2001relation}. In map-matching evaluation, we transform the Average Hamming Distance to calculate the model accuracy. The equation is shown as,

\begin{linenomath}
  \begin{equation}
  AHD_{point/segment} = \frac{\sum {\#\,of\,matched\,points/segments}}{\sum {\#\,of\, points/segments}}
  \end{equation}
\end{linenomath}

However, in segment-level matching, the length of the result and ground-truth sequences are not always identical. For example, the result sequence is $[7,8,9]$, whereas the true sequence can be $[7,8,9,10]$. Therefore, to calculate the accuracy at the segment level, it is necessary to implement sequence alignment algorithms. In this paper, the Needleman-Wunsch algorithm, which is an algorithm used in bioinformatics to align protein or nucleotide sequences is adopted to balance the length of the result and ground-truth segment sequences \cite{likic2008needleman}. 

\paragraph{\textbf{F-Score}}
The second performance evaluation metric is the F-score, which is one of the most commonly used measurements of a model’s accuracy \cite{sokolova2006beyond}. It is used to evaluate binary classification systems, which classify examples into ‘positive’ or ‘negative’. The traditional F-score (or $F_1 score$) uses the harmonic mean of precision and recall of the model. The related equation is shown as, 

\begin{linenomath}
  \begin{equation}
  F_{score} = \frac{2}{\frac{1}{recall}\cdot\frac{1}{precision}} = 2 \times \frac{precision \times recall}{precision + recall}
  \end{equation}
\end{linenomath}

\paragraph{\textbf{BLEU score}} 
BLEU (Bilingual Evaluation Understudy) score \citep{papineni2002bleu} is one of the most commonly used metrics in machine translation and sequence to sequence learning problems. Recently, BLEU is applied as a measurement of effectiveness in trajectory-based researches \citep{choi2021trajgail,sun2021joint}. In machine translation, BLEU scans the reference sentences to see if the translated sentences include the same words or contiguous sequence of $n$ elements. BLEU uses a modified form of precision to compare reference sequences and a candidate output sequence by using the clipping method. The number of each chunk is clipped to a maximum count $(m_{max})$ to avoid generating the same chunks to get a high score in the output sequence. The equation of modified precision is shown as,  

\begin{linenomath}
  \begin{equation}
  P_n= \frac{\sum\limits_{i\in C} min(m_i\, m_{i,\max})}{w_t}
  \end{equation}
\end{linenomath}

\noindent
where $n$ is the number of elements considered as chunk; $C$ is a set of unique chunks in the output sequence; $m_i$ is the number of occurrences of chunk $I$ in the output; $m_(i, max)$  is the maximum number of occurrences of chunk $i$ in one of the reference sequences; and $w_t$ is the total number of chunks in the output sequence. 

$BLEU_n$ score consists of the geometric mean of $P_n$ and a term of brevity penalty. The brevity penalty is used to prevent short candidates from getting high scores. The $BLEU_n$ is shown as,

\begin{linenomath}
  \begin{equation}
  BLEU_n= min(1,\frac{length_{gen}}{length_{ref \,, close}})(\prod \limits^{n}_{i=1}P_{i})^{\frac{1}{n}}
  \end{equation}
\end{linenomath}

\noindent
where $length_{gen}$ represents the length of the output sequence; $length_{ref, close}$ is the length of a reference sequence that has the closest length to the output sequence. We use $n = 3$ for evaluation because the minimum length of the sequence is 3 in the results.

\subsection{Result}\label{sec:result}

The results of performance evaluations for the proposed map-matching model are discussed in four different perspectives: \textbf{\emph{1) Performance evaluation at pre-training stage, 2) Performance evaluation at fine-tuning stage, 3) Analysis on trajectory-level performance improvement, and 4) Attention on attention mechanism in Transformer.}}
%
%
In Section~\ref{sec:pretraining evaluation}, we use generated trajectory data to pre-train the deep learning models and evaluate the matching performance using ground-truth testing data. 
We generate two trajectory datasets that have different noise standard deviations($\sigma_{noise}$) to evaluate the noise effects on model performance. The first dataset is built without noise, while the second dataset is constructed with Gaussian distributed noise with the $15m$ standard deviation for each coordinate.
%
%
In Section~\ref{sec:fine-tuning evaluation}, we fine-tune two pre-trained Transformer models to see how much the method improves matching performances. As stated earlier, we randomly select $931$ trajectories (70\% of ground-truth data) for fine-tuning the model and the last 400 trajectories are used in model evaluation. Figure ~\ref{fig:Transformer} shows four components, which are represented by red circles. The full encoder modules, for example, are used to extract features of input data, whereas the full decoder modules are used to convert the extracted information from the encoder modules to the target domain. We gradually increase the number of components for fine-tuning in order to find the best scenarios for our proposed map-matching model.
At analyzing stage, we primary analyze the result at the trajectory level in Section~\ref{sec:trajectory analysis} to reveal how the fine-tuning approach improves the map-matching performances and reduces the real-to-virtual gaps when compared to the pre-training model at trajectory level.
Finally, in Section~\ref{sec:weight analysis} attention mechanisms of fine-tuned Transformer model are analyzed in order to better understand the map-matching mechanisms of our proposed model.

\subsubsection{Performance Evaluation at Pre-training Stage}\label{sec:pretraining evaluation}
In trajectory generation for model pre-training, we assume the spatial noise at longitude and latitude follows zero-mean Gaussian distribution, as stated in GPS trajectory generation in \ref{sec:transfer learning}, To make more realistic trajectories, we choose $15m$ as the noise standard deviation ($\sigma_{noise}$) since the standard deviations of the longitude and latitude noise $15.653m$ and $14.153 m$, respectively.
We evaluate our model with four baseline models on the ground-truth test dataset and corresponding results are shown in Table~\ref{tab:pre_trained point} and Table~\ref{tab:pre_trained segment}.
FMM and ST-Matching models do not require training procedures since they are rule-based models, and the performance of the model is evaluated based on the test dataset with 400 trajectories.
Three deep learning models (LSTM-based seq2seq, LSTM-based attentional seq2seq and Transformer models) are trained on three different datasets: 1) ground-truth training dataset with 931 trajectories ($D_{GT}$), 2)generated trajectory without noise ($\sigma_{noise}$ = 0\emph{m}) ($D_{Gen,0m}$) and generated trajectory with Gaussian distributed noise ($\sigma_{noise}$ = 15\emph{m}) ($D_{Gen,15m}$). Then, the performance of each model is evaluated based on the test dataset with 400 trajectories.     
%
%

From the perspective of training data, the results reveal that $D_{GT}$ is not enough to build a high-performing map matching model because it doesn't cover the whole target area and various scenarios. In addition, even if the models trained by $D_{Gen,0m}$ outperform the models trained by $D_{GT}$, the performance of the models can improve more with $D_{Gen,15m}$. The difference comes from the existence of Gaussian distributed noise in generated trajectory data. The $D_{Gen,15m}$ is more close to real-world trajectory since the noise distribution used in $D_{Gen,15m}$ is based on real ground-truth data. On the contrary, $D_{Gen,0m}$ doesn't add noise which has the difference between real-world trajectories. The result shows the potential of using generated data to overcome the lack of labeled data problems at the training stage.
%
From the perspective map-matching model, two rule-based models (FMM and ST-matching) achieve similar performance in three metrics with two levels. The three deep learning models trained by generated trajectories ($D_{Gen,0m}$, $D_{Gen,15m}$) show better matching performances than two rule-based models. The transformer-based map-matching model outperforms the other deep learning models. Therefore, we can conclude that our proposed Transformer-based map-matching model trained by generated trajectories with Gaussian distributed noise outperforms baseline models in the map-matching process.
From the perspective of model evaluation, the rule-based map-matching models perform better at point-level matching than segment-level matching. The reason behind this is that for the result that when the route integrity is broken, the metrics at point level show the higher scores at segment level matching. These cases commonly happen in confusing regions such as the initial segment, last segment, and the transition area between two consecutive segments. On the contrary, in deep learning-based models, the overall performances are better in segment-level results, indicating that deep-learning models are better to extract drivers' routes and are ideal for developing map-matching models.      
%

\begin{table}[!ht]
	\caption{Pre-trained model performance at point level}\label{tab:pre_trained point}
	\begin{center}
		\begin{tabular}{p{1.5cm} p{3cm} p{4cm} p{1.5cm} p{1.5cm} p{1.5cm} }
	Type	&	Model             & Data           & AHD     & F-score & BLEU \\\hline\hline
    \multirow{11}{*}{Point}&\multirow{1}{*}{ST-Matching} 
    &  -  & 0.8719  & 0.7151  & 0.8548 \\\cline{2-6}
			&\multirow{1}{*}{FMM}         
	& -  & 0.8573  & 0.7184  & 0.8517 \\\cline{2-6}
			&\multirow{3}{*}{LSTM EncDec}   
	& ground-truth training data & 0.4829  & 0.348  & 0.5679   \\
	\cline{3-6}        &  & generated data ($\sigma_{noise}=0m$)  & 0.6340  & 0.4933  & 0.7032 \\
	\cline{3-6}        &  & generated data ($\sigma_{noise}=15m$)  & 0.8960  & 0.8386  & 0.9029 \\
	\cline{2-6}
			
	&\multirow{3}{*}{LSTM Attn EncDec} 
	& ground-truth training data & 0.6049  & 0.4475  & 0.7150  \\
	\cline{3-6} & & generated data ($\sigma_{noise}=0m$) & 0.8018  & 0.7367  & 0.9004 \\
	\cline{3-6} & & generated data ($\sigma_{noise}=15m$) & 0.9563  & 0.9372  & 0.9557 \\
	\cline{2-6}
	&\multirow{3}{*}{Transformer} 
	& ground-truth training data & 0.7644  & 0.6799  & 0.6585 \\
	\cline{3-6} & & generated data ($\sigma_{noise}=0m$)   & 0.8623  & 0.7860  & 0.8905\\
	\cline{3-6} & & generated data ($\sigma_{noise}=15m$) & 0.9737  & 0.9558  & 0.9724 \\
	\hline\hline
		\end{tabular}
	\end{center}
\end{table}

\begin{table}[!ht]
	\caption{Pre-trained model performance at segment level}\label{tab:pre_trained segment}
	\begin{center}
		\begin{tabular}{p{1.5cm} p{3cm} p{4cm} p{1.5cm} p{1.5cm} p{1.5cm} }
	Type	&	Model             & Data           & AHD     & F-score & BLEU \\\hline\hline	
	
	\multirow{11}{*}{Segment}&\multirow{1}{*}{ST-Matching} 
	&  - & 0.7408  & 0.6915  & 0.6668 \\\cline{2-6}
    &\multirow{1}{*}{FMM} & -         & 0.7477  & 0.6950  & 0.6775 \\
	\cline{2-6}
    &\multirow{3}{*}{LSTM EncDec}
    & ground-truth training data & 0.5714  & 0.4739  & 0.4546   \\
	\cline{3-6} && generated data ($\sigma_{noise}=0m$) & 0.7127  & 0.6004  & 0.6417 \\
	\cline{3-6} && generated data ($\sigma_{noise}=15m$) &0.9156  & 0.8852  & 0.9209 \\ 
	\cline{2-6}		
	&\multirow{3}{*}{LSTM Attn EncDec}
	& ground-truth training data & 0.7007  & 0.5947  & 0.6272   \\
	\cline{3-6}&& generated data ($\sigma_{noise}=0m$)    &0.9027  & 0.8478  & 0.8743 \\
	\cline{3-6}&& generated data ($\sigma_{noise}=15m$)    &0.9741  & 0.9601  & 0.9660 \\
	\cline{2-6}		
	&\multirow{3}{*}{Transformer}
	& ground-truth training data & 0.7068  & 0.5816  & 0.7423 \\
	\cline{3-6} & & generated data ($\sigma_{noise}=0m$) & 0.9150  & 0.8633  & 0.9020\\
	\cline{3-6} & & generated data ($\sigma_{noise}=15m$) & 0.9784  & 0.9643  & 0.9751 \\
	\hline\hline

		\end{tabular}
	\end{center}
\end{table}

\clearpage

\subsubsection{Performance Evaluation at Fine-tuning Stage} \label{sec:fine-tuning evaluation}
In the following series of experiments, we study how the real-to-virtual gaps between generated trajectories and the real ground-truth trajectories are reduced by using the fine-tuning method. From the previous section, we conclude that Transformer-based map-matching models show better performances among three deep learning models with three different datasets. As a result, we choose to fine-tune the Transformer-based models to get high-performing map-matching models.

In this experiment, two pre-trained Transformer models, which are trained with $D_{Gen,0m}$ and $D_{Gen,15m}$, are chosen to do fine-tuning with $D_{GT}$.The components of Transformer for fine-tuning are depicted in Figure~\ref{fig:Transformer}: {\large \textcircled{\small 1}} output module, {\large \textcircled{\small 2}} normalization layers in encoder and decoder modules,{\large \textcircled{\small 3}} full encoder modules, and {\large \textcircled{\small 4}} full decoder modules. We gradually increase the number of tuning components from the output module to the entire model to analyze the effects of fine-tuning on model performances. We fine-tune the models 30 times and the averaged results are shown in Table~\ref{tab:fine-tuning 0m} and \ref{tab:fine-tuning 15m}.

Table~\ref{tab:fine-tuning 0m} shows the fine-tuning results at $D_{Gen,0m}$-based pre-trained Transformer model. From the perspective of model performance, overall performance is improved after fine-tuning at both point and segment levels. To be more specific, the pre-trained model performs poorly when compared to the pre-trained model trained with $D_{Gen,15m}$ since the characteristics of the pre-trained dataset are different from the real ground-truth dataset. From the perspective of fine-tuning components, there are no significant performance differences except tuning the output layers. This case can be treated as scenario 3, where the size of the ground-truth dataset is small and different from the pre-trained dataset as stated in Section~\ref{sec:transfer learning}. In this case, it is not enough to only fine-tuning the output module to get a high-performing model. Instead, We must find appropriate fine-tuning modules to improve model performance. In this instance, fine-tuning is used to improve model performance at both point and segment levels by reducing real-to-virtual gaps.

Table~\ref{tab:fine-tuning 15m} represents the fine-tuning results of the pre-trained model trained with $D_{Gen,15m}$. From the perspective of model performance, in the point-level result, there are minor improvements. In the segment-level result, although the improvements are small, they are noticeable when compare to the point-level findings. Also, there are no significant variations between the results of each fine-tuned component. One of the key explanations of these findings is that the pre-trained dataset exhibits similar characteristics to the real ground-truth dataset. Specifically, the pre-trained model shows high performance because the generated trajectories used in model training are close to real-world trajectories. Therefore, the fine-tuned results are not obvious at each component. We can consider this case as scenario 4, where the size of the ground-truth dataset is small but similar to the pre-trained dataset, as stated in Section~\ref{sec:transfer learning}. According to the results, the fine-tuning method in this case is prominent in improving segment-level matching performance. In the next Section~\ref{sec:trajectory analysis} we will discuss the findings in detail. 

There are two main conclusions throughout the results. First, the fine-tuning method is beneficial in improving model performance. Even though there are performance differences between two fine-tuned pre-trained models, both of them show high performance in the map-matching task. Especially, the pre-trained model used in Table~\ref{tab:fine-tuning 0m} becomes a high-performing model after using the fine-tuning method. This finding shows the potential of the fine-tuning method in building a more general map-matching model with simply generated trajectories. In practice, it is difficult to generate trajectories that are close to target real-world trajectories precisely without prior information. As a result, it is important to develop a more general map-matching model that not only shows high performance but also can be built without any prior information. 
Second, the prior information of real-world trajectories is efficient in model development. The efficiency is demonstrated in fine-tuning process. As previously stated, $D_{Gen,15m}$ are close to real-world trajectories so that the pre-trained model outperforms the others. As a result, it is considerably easier to find suitable components for fine-tuning the model when compared to the pre-trained model, which is shown in Table~\ref{tab:fine-tuning 15m}. In other words, if we generate datasets using prior information, we can reduce the costs of finding appropriate fine-tunable components. 

\clearpage

\begin{table}[!ht]
	\caption{Matching results of fine-tuned pre-trained model with $D_{Gen,0m}$ (bar graph scaled from 0.75 to 1) }\label{tab:fine-tuning 0m}
\includegraphics[width=\linewidth]{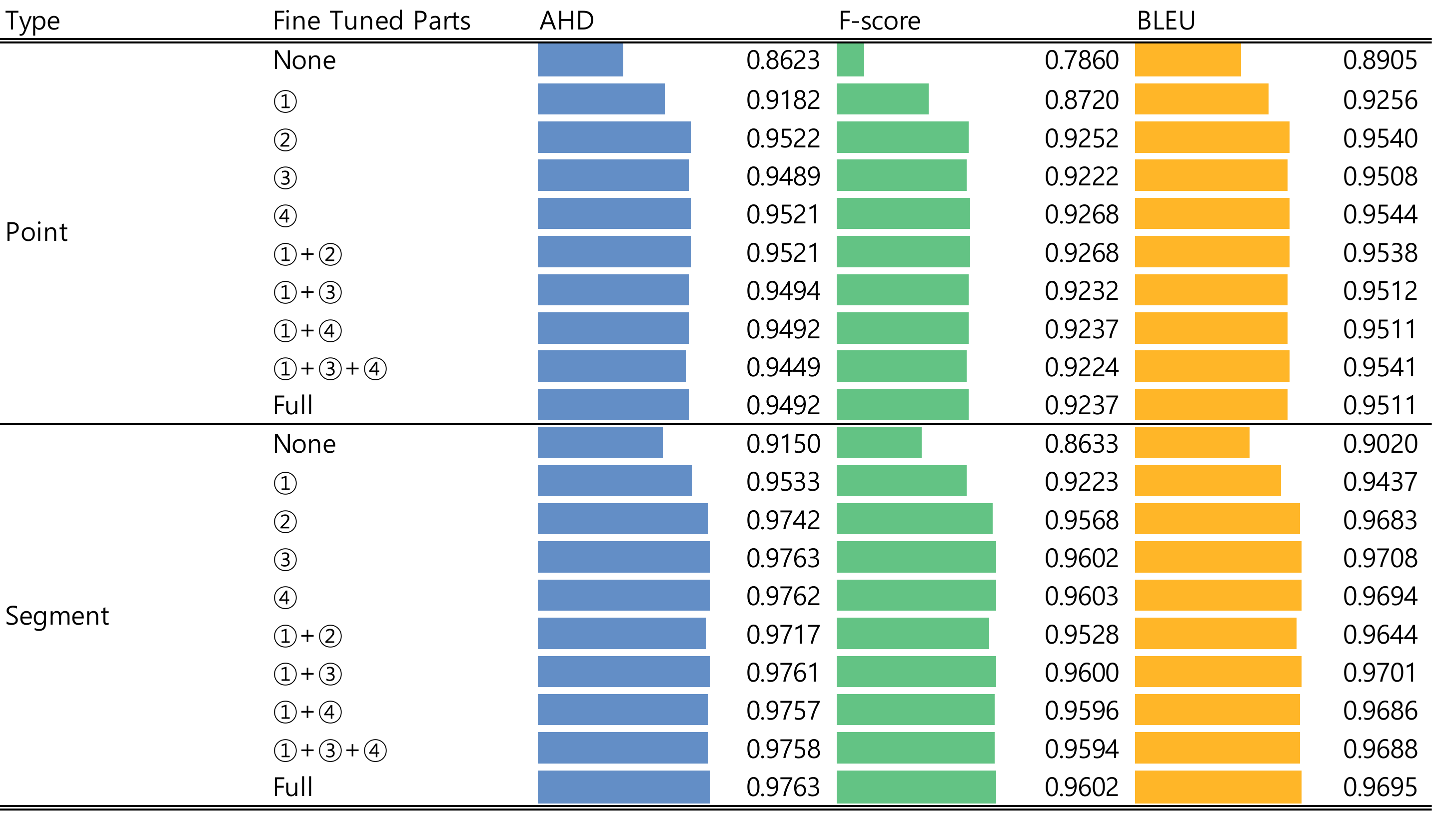}
\end{table}

\begin{table}[!ht]
	\caption{Matching results of fine-tuned pre-trained model with $D_{Gen,15m}$  (bar graph scaled from 0.9 to 1) }\label{tab:fine-tuning 15m}
	\includegraphics[width=\linewidth]{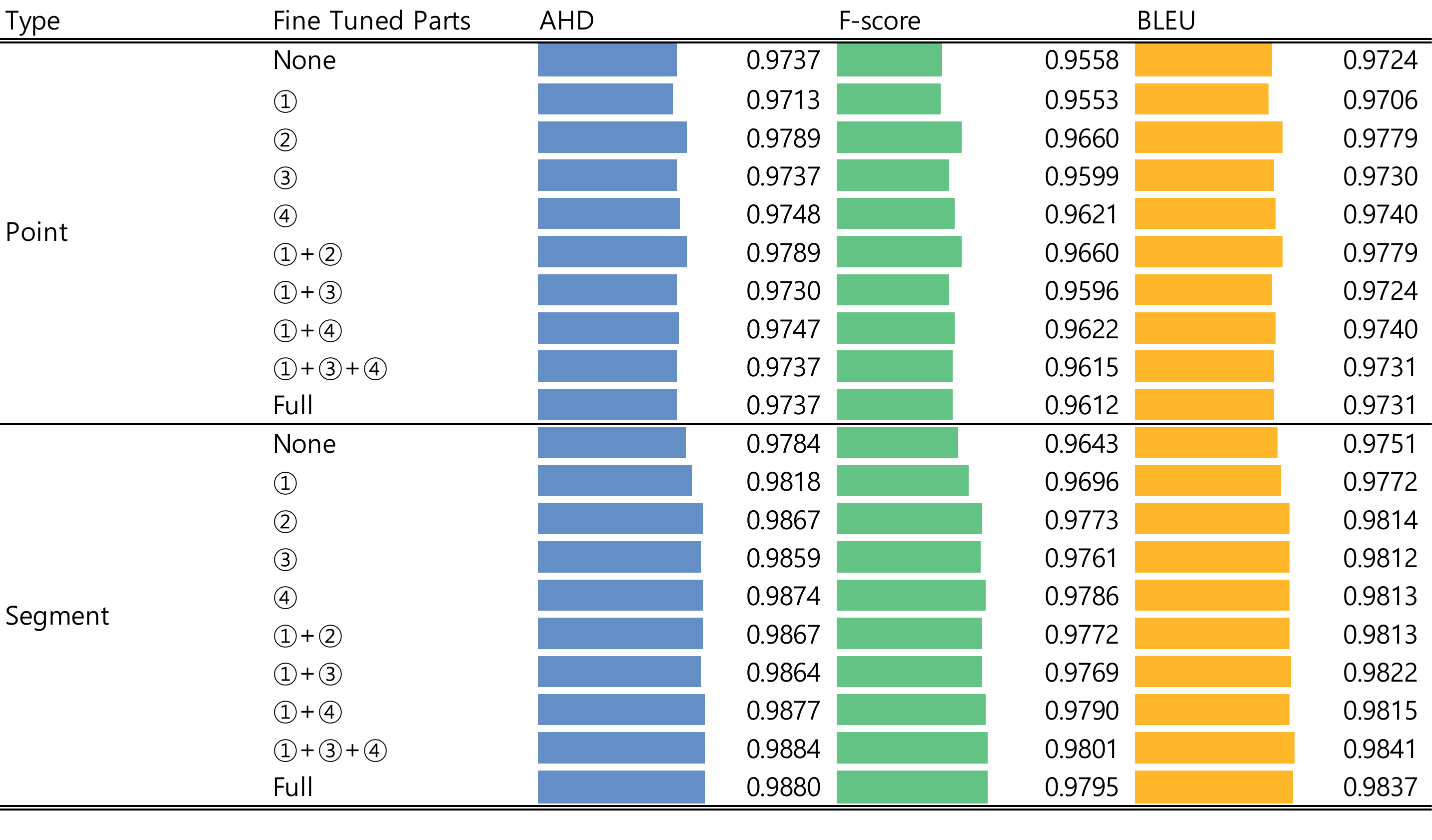}
\end{table}

\clearpage

\subsubsection{Analysis on Trajectory-level Performance Improvement } \label{sec:trajectory analysis}

\begin{figure}[!ht]
  \centering 
  \includegraphics[width=1\textwidth]{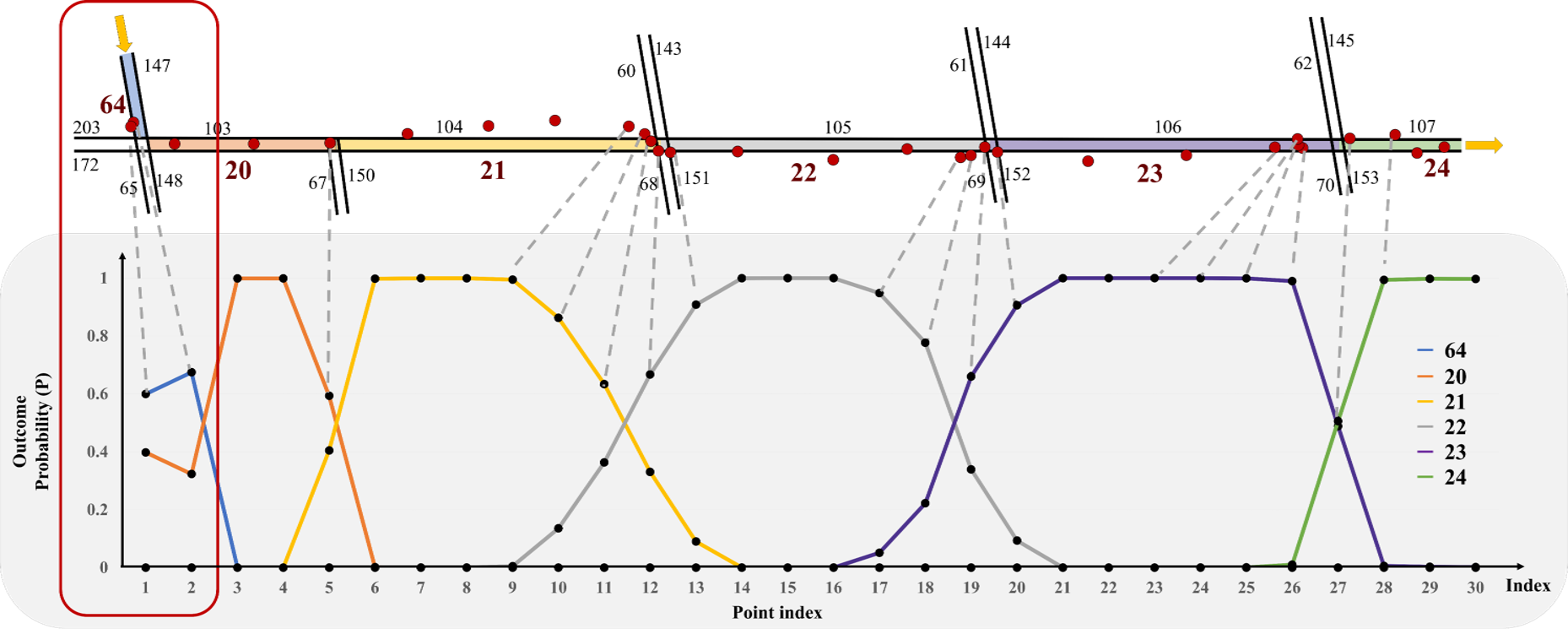}
  \caption{The matching result for fine-tuned Transformer model}\label{fig:Fine-tuned result}
\end{figure}

In this section, we analyze the result at the trajectory level and investigate how the fine-tuning approach improves the matching performance and reduces the real-to-virtual gaps between generated and real-world trajectories. 
Figure~\ref{fig:Fine-tuned result} depicts one of the matching results of the fine-tuned map-matching model which shows the best performance from the previous section. This figure is divided into two parts: the upper portion shows road network scenarios, while the lower part of the figure presents the probability of GPS points matching at each segment. In the upper part of the figure, the red dots represent a vehicle's moving positions on the road, which are randomly distributed. The colored link represents the ground-truth segment, where each one has its color. The ground-truth segment-level route is $[64,20,21,22,23,24]$. We calculate the probability of each moving point matching on the ground-truth segment and plot the associated result on the lower part of Figure~\ref{fig:Fine-tuned result}. 
From the figure, we find that the fine-tuned model matches all the points and segments appropriately. In the figure, the matching probability of each GPS point on the corresponding segment is close to 1 instead of the confusing region such as transition area between two segments, first and last segment. Furthermore, our fine-tuned model performs well in the transition region. For example, the fifth GPS point is recorded between segments 20 and 21. The summation of probability matching on two segments is close to 1, indicating that the map-matching model performs well.

\begin{figure}[!ht]
  \centering
  \includegraphics[width=0.8\textwidth]{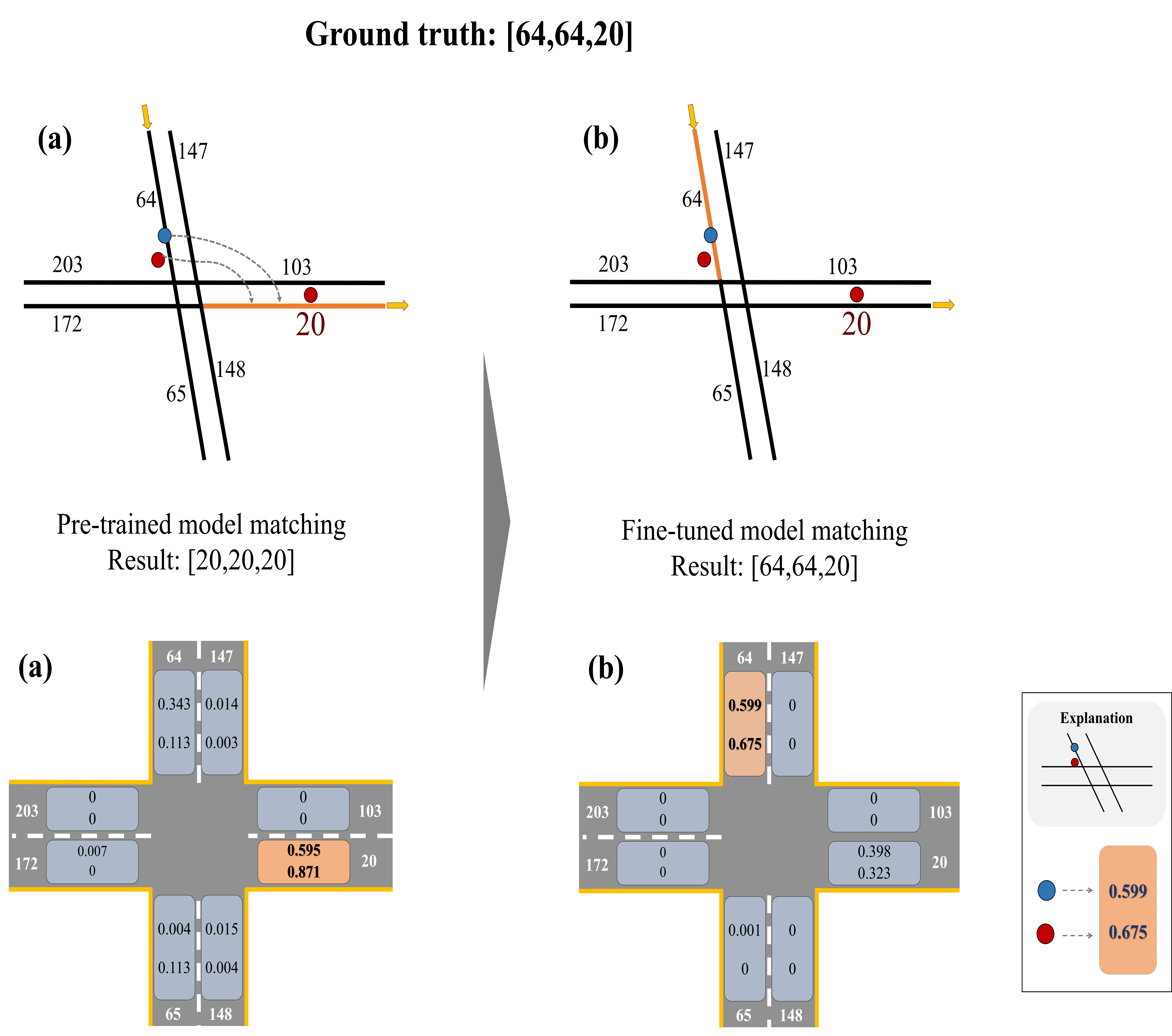}
  \caption{Amplified results for the result figure}\label{fig:Amplified_results}
\end{figure}

However, if we test the same scenario on the pre-trained Transformer model, the model can not match the first two points on segment 64. To compare the matching outcomes from pre-trained and fine-tuned models in detail, the probability results for the first two GPS points are shown in Figure~\ref{fig:Amplified_results}. 
In Figure~\ref{fig:Amplified_results}, the ground-truth point-level route is $[64,64,20]$ for the first three points. In contrast, the matching result is $[20,20,20]$  with a high matching probability in the pre-trained model. There is a more than 50\% chance of matching the first two points at segment 20. 
One of the reasons behind this is the pre-trained model considers the first two points as the noise points which are generated at segment 64. To calibrate the aforementioned problems, the fine-tuning method is applied and the result sequence is turned to $[64,64,20]$. The probability for matching the first two points at segment 64 is improved 25.6\% and 56.2\%, respectively. Therefore, we can conclude that the fine-tuning method effectively reduces the real-to-virtual gaps between generated and real-world trajectories. This method the model more robust to complex real-world scenarios and produces much more realistic results.


\clearpage


\subsubsection{Analysis on Attention Mechanism in Transformer} \label{sec:weight analysis}
In this section, we use attention mechanisms to analyze the map-matching process, which shows how the model considers the internal correlation of GPS points and matches the road segments throughout the input GPS trajectories. We extract the attention weights from the encoder and decoder layers to investigate the correlation between GPS points and the relationship between GPS points and road segments, respectively.

\begin{figure}[!ht]
  \centering 
  \includegraphics[width=0.8\textwidth]{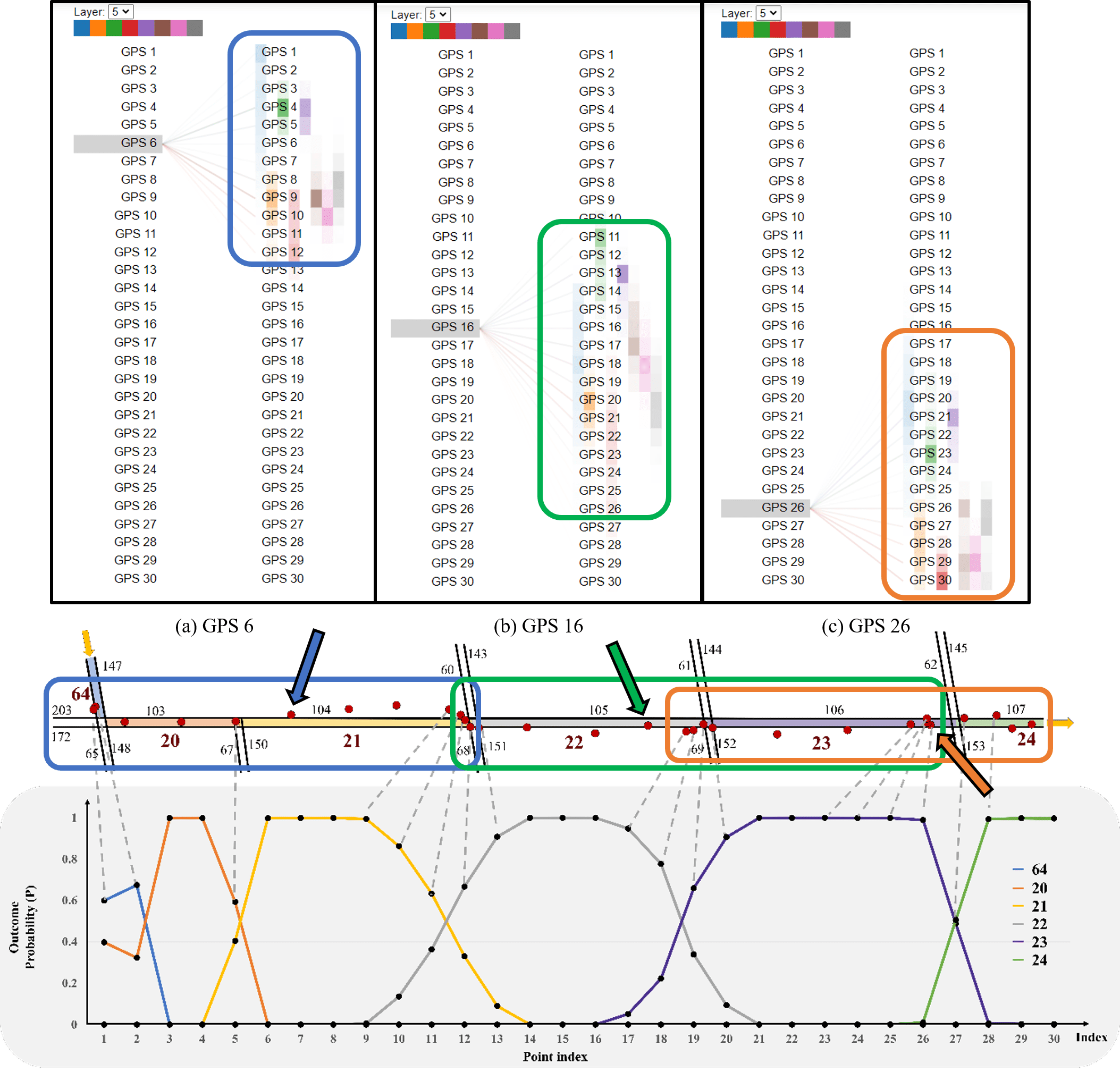} 
  \caption{Internal correlation of GPS points in the trajectory (a) correlated points of GPS6 (b) correlated points of GPS16 (c) correlated points of GPS26}\label{fig:encoder_attention}
\end{figure}

In encoder attention weight analysis, We use the visualization tool provided by \cite{vig-2019-multiscale} to clearly see the correlations, and the result is shown in Figure~\ref{fig:encoder_attention}. The upper part of the figure depicts the internal correlation of GPS points in the trajectory. We choose GPS 6, GPS 16, and GPS 26 as test points and see how the related points are changed in different GPS points. There are eight colors that represent the specific weight of each head. Here, the darker hue means a stronger correlation with the target point. The result shows that there is a certain range of GPS points that have a strong correlation with target points. Furthermore, each GPS point has its own related range. We determine the three related ranges from the upper figure and plot them in the lower figure to find the position of correlated points on the road segment in detail. In Figure, the colored boxes represent the related ranges. According to the figure, the GPS points, which are on the current and adjacent road segment, have a strong correlation with the target point. For example, in Figure~\ref{fig:encoder_attention} (b), the GPS16 has a correlation with the points from GPS11 to GPS26 and the green box stretches from the end of segment 21 to the end of segment 23.

\begin{figure}[!ht]
  \centering 
  \includegraphics[width=0.8\textwidth]{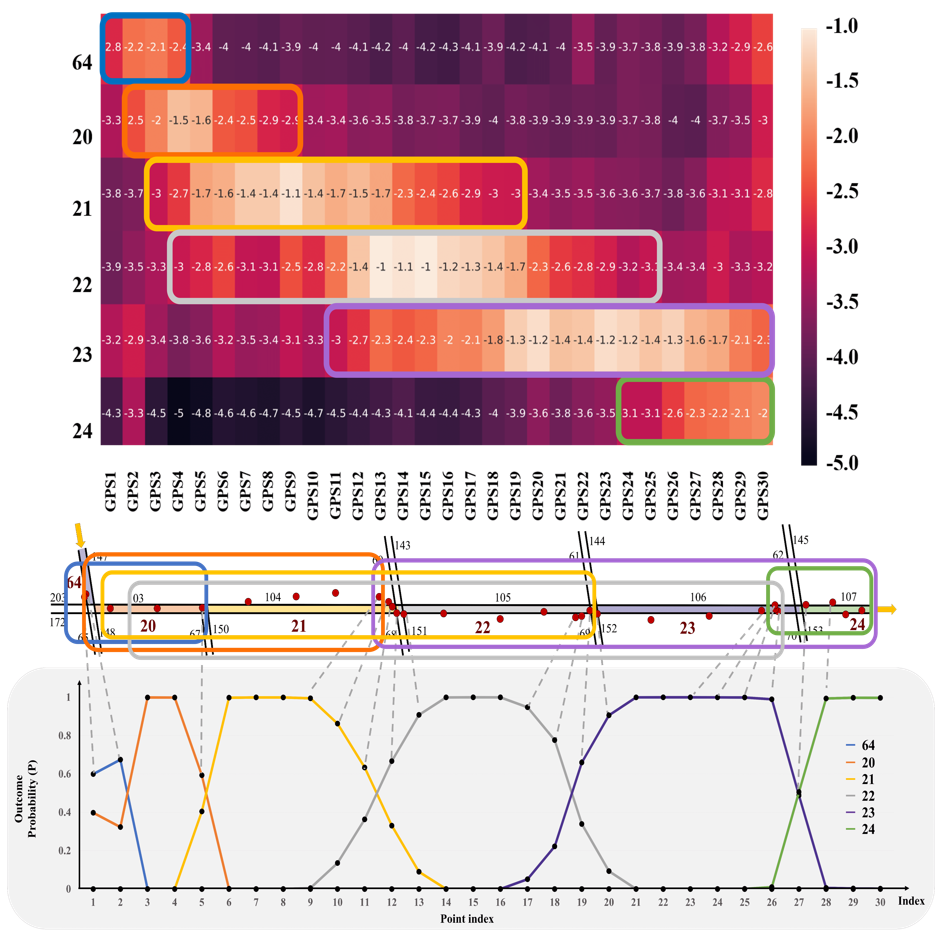} 
  \caption{Plot of the decoder attention mechanisms}\label{fig:decoder_attention}
\end{figure}

In decoder attention analysis, the logarithm transformation of the weights is used to plot the figure since the value of the weights are too small to depict in detail. The bright part in the figure denotes large attention weights, which can also be interpreted as a high correlation with decoding results. From Figure~\ref{fig:decoder_attention}, we find that specific ranges of points have high attention weights. To determine the range, we use the threshold value of -3.15, which is the average mean and median of total log weights. We determine the range of GPS points that affect the matching results in decoder modules and depict them in Figure~\ref{fig:decoder_attention}. The stated ranges are represented as colored boxes. According to the determined ranges, we discover that in order to match the current road segment, not only GPS points on the current road segment but also GPS points on neighboring segments are involved in the matching process, which has a similar result in encoder weight analysis. For example, the yellow box in Figure~\ref{fig:decoder_attention} illustrates the range which influences the determination of segment 21. The box stretches from third to nineteenth points where they are on the segment 20,21 and 22. The result shows that while identifying segment 22, points on segments 20 and 22 have a substantial influence on the decoding process. Furthermore, since there is only one segment adjacent to the first and last segments, the points from one road segment are used in the decoding process, implying that there is less information in the matching process compared to other road segment matching. As a result, there are several matching errors in the first and end segment matching process. 

\newpage

\section{Conclusion} \label{sec:conclusion}
This study proposes a framework for developing a novel deep learning-based map-matching model in the limited ground-truth data environment. In the proposed map-matching model, an advanced deep-learning model \textit{Transformer} is used to improve model accuracy by capturing the internal correlation of input GPS points and the external relationship between input and output. To solve the data sparsity problem in model training, a \textit{Transfer-learning approach}, which pre-trains the model with generated data and fine-tunes the pre-trained model with real ground-truth data is applied. 

In terms of contributions, this study has made several improvements in the field of deep learning-based map-matching models. The proposed model improves matching accuracy by using the Transformer model, which considers the internal correlation of GPS points and the external relationship between input trajectory and output segments. This overcomes the disadvantages of the previous deep learning-based map-matching model that they can not take into account the data relationships. In addition, to solve the data sparsity problem for developing a high-performing map-matching model, the transfer learning approach is adopted in this study. Specifically, a large number of trajectories are generated based on road network information to pre-train the model, and then a limited amount of available ground-truth data is used to fine-tune the model to reduce the real-to-virtual gaps. The proposed model shows the possibility of using generated trajectories to solve the map-matching problems in the urban environment. This overcomes the problem of developing map-matching models not having enough ground-truth data. We also analyze the matching mechanisms of the Transformer in the map-matching process, which helps to interpret the input data internal correlation and external relation between input data and matching results. Finally, we consider the map-matching task from the data perspective and propose three related metrics at point and segment levels, which help in developing more high-performing map-matching models.

The model's performance is evaluated on the Gangnam DTG dataset, which contains moving patterns of taxis in the Gangnam district. The performance evaluation is divided into two levels: point-level evaluation and segment-level evaluation. The point-level evaluation mainly focuses on how the model matches each point to its corresponding segment correctly. Conversely, in the segment-level evaluation, the main concern is how the model matches the integrated route (or segment-level route) correctly. Both levels are evaluated in terms of AHM, F-score, and BLEU, which are widely used metrics in sequence modeling. At the pre-training stage, the results show that the generated data-based pre-trained models show better performance than rule-based models (FMM and ST-matching). In addition, the Transformer-based map-matching model outperforms other deep learning-based models (LSTM-based seq2seq and LSTM-based attentional seq2seq models) in three different datasets. At the fine-tuning stage, we fine-tune the two pre-trained Transformer-based map-matching models trained by different generated datasets. The results indicate that fine-tuning method can reduce the real-to-virtual gaps in both models. Specifically, in the pre-training model which shows better performance, fine-tuning is principal to make the results more realistic. In the other pre-training model, fine-tuning is mainly used to improve model performance. To further analyze the results that how the fine-tuning method makes the result more realistic, we choose one noisy trajectory as an example. The results show that the pre-trained model can match the first two points correctly after fine-tuning. In addition, we also analyze the attention mechanisms to find the internal correlation of GPS points and the external relationship between input trajectory and output results. The results show that the points which are on the current and adjacent road segment have a strong correlation with the target point. In addition, similar to the previous findings, while identifying a certain segment, not only the points on the target segment but also the points on the neighboring segments have correlations.

There are several directions in which the current study can be extended to further improve the map-matching performance. Currently, we use a simple rule to generate the trajectories for pre-training. There are, however, other variables that can help improve matching performance in addition to the proposed method. For instance, traffic volumes for each road segment, traffic signal and road geometry information can all provide additional information to further improve the quality of generated data. In addition, it is difficult to identify the relationship between the model performance and the number of real ground-truth data used in fine-tuning due to the lack of real ground-truth trajectories. Furthermore, since our proposed model is an offline map-matching model that is only used as a preprocessing step for trajectory-based applications, we should develop an online deep-learning based model that can be applied in real-time service. While the current study focuses on demonstrating the potential of using generated data with a transfer-learning approach, we will consider incorporating other variables in trajectory generation and using more real ground-truth trajectories to further reduce the virtual-to-real gaps and apply the system to a real-time map-matching service in future work.

\newpage
\printcredits

\bibliographystyle{cas-model2-names}
\bibliography{partc}

\end{document}